\documentclass{article}

\usepackage[english]{babel}

\usepackage[letterpaper,top=2cm,bottom=2cm,left=3cm,right=3cm,marginparwidth=1.75cm]{geometry}
 
\usepackage{amsfonts} 
\usepackage{amssymb}
\usepackage{amsmath}
\usepackage{graphicx}
\usepackage[colorlinks=true, allcolors=blue]{hyperref}
\usepackage{subfigure}
\usepackage{lscape} 
\usepackage{mdframed} % This one is new!!!
\usepackage{adjustbox}
\usepackage[ruled,vlined]{algorithm2e}
\usepackage{todonotes} 
\usepackage{amsthm}

\theoremstyle{definition}
\newtheorem{example}{Example}%
\newtheorem{remark}{Remark}%
\newtheorem{assumption}{Assumption} 

\newcommand{\drugs}{d}   
\newcommand{\targets}{t}   
\newcommand{\preds}{p}    
   
\newcommand{\duals}{a}   
\newcommand{\ds}{{\mathcal{D}}}   
\newcommand{\ts}{{\mathcal{T}}}

\newcommand{\barr}{{\bar{r}}}
\newcommand{\bars}{{\bar{s}}}
\newcommand{\bbarr}{{\bar{r}}}
\newcommand{\bbars}{{\bar{s}}}
\newcommand{\barm}{{\bar{m}}}
\newcommand{\barn}{{\bar{n}}}
\newcommand{\barq}{{\bar{q}}}

\newcommand{\x}{x}            % vector x bolded
\newcommand{\y}{y}            % vector y bolded
\newcommand{\bs}{{s}}           % vector s bolded
\newcommand{\br}{{r}}           % vector s bolded
\newcommand{\Jobj}{{J}}              % Objective function

\newcommand{\bxi}{\xi}          % vector \xi bolded
\newcommand{\bzero}{{0}}          % vector 0 bolded
\newcommand{\batch}{\mathcal{B}}          % vector 0 bolded
\newcommand{\iLMBM}{{\sc InexactLMBM }}
\newcommand{\iLMBMNS}{{\sc InexactLMBM}}
\newcommand{\lmbm}{{\sc LMBM }}

\newcommand{\slmbmns}{{StoILMBM}}
\newcommand{\slmbm}{{StoILMBM }}

\newcommand{\sgvt}{{\sc sGVT }}
\newcommand{\sgvtns}{{\sc sGVT}}
\newcommand{\spkl}{{\sc SPaiK }}
\newcommand{\spklns}{{\sc SPaiK}}
\newcommand{\R}{\mathbb{R}}
\newcommand{\N}{\mathbb{N}}
\newcommand{\norm}[1]{\lVert#1\rVert}
\newcommand{\ignore}[1]{} % 

\title{Scalable Pairwise Kernel Learning 
       with Stochastic Vec Trick
       }
\author{Napsu Karmitsa\footnote{\texttt{napsu@karmitsa.fi}}, Tapio Pahikkala, Antti Airola\\
{\small Department of Computing, University of Turku, FI-20014 Turku, Finland}}

\begin{document}
\maketitle

\begin{abstract}
\noindent
Pairwise learning is a specialized form of supervised learning that focuses on predicting outcomes for pairs of objects.
In this work, we introduce \spklns, a new scalable kernel learning method tailored for pairwise settings. Our approach preserves the expressive power of kernel methods while substantially reducing computational and memory requirements. The key innovation is the {\em stochastic generalized vec trick} (sGVT), a stochastic extension of the sparse Kronecker product multiplication algorithm, which enables efficient large-scale training with pairwise kernels. By incorporating sGVT, \spkl\ makes it possible to apply kernel-based pairwise learning to datasets of a size previously out of reach.
We evaluate the performance of \spkl\ on seven real-world drug-target affinity datasets and compare the results with state-of-the-art methods in pairwise learning.
\smallskip

\end{abstract}

\section{Introduction}
\begin{figure}[ht] 
%\phantom{M}
\begin{center}
\includegraphics[width=0.690\linewidth, trim= .01 .01 .01 .01, clip=true]{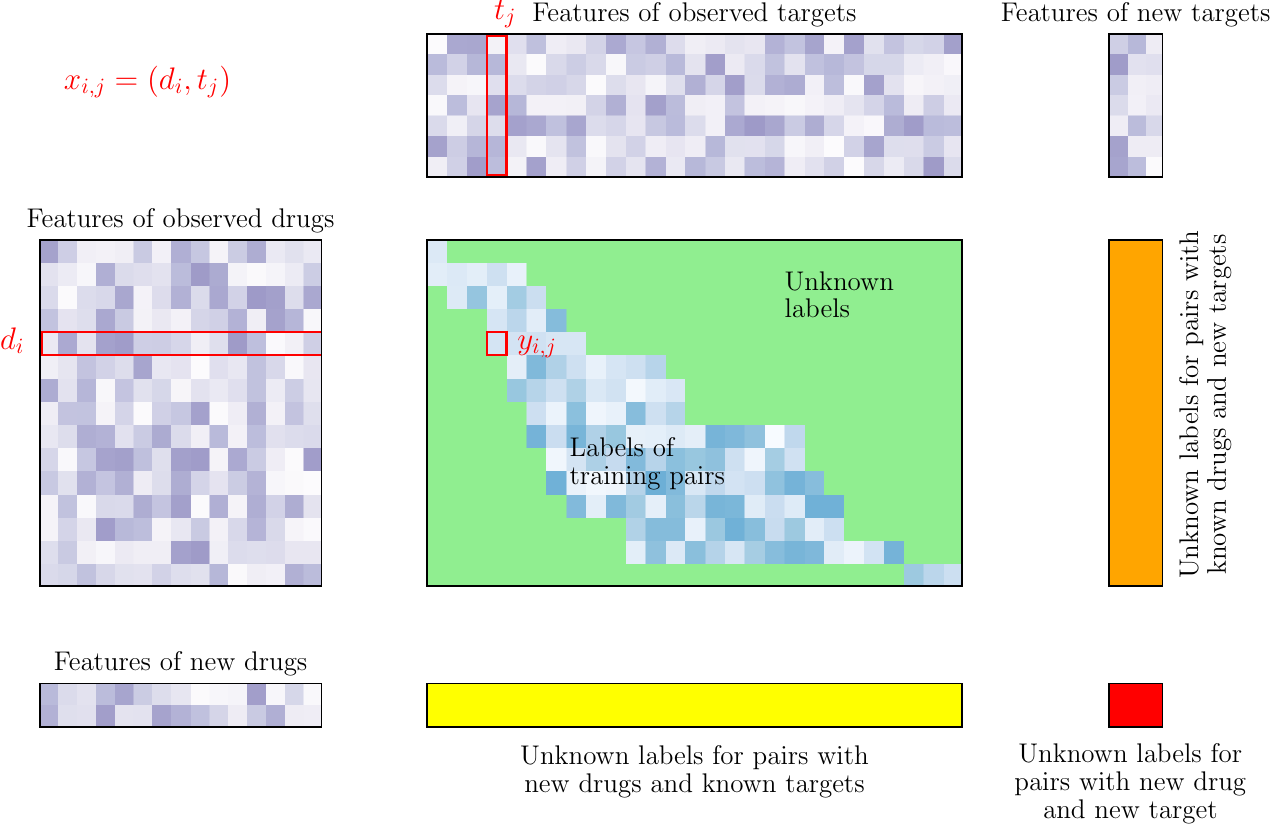}    
\end{center}
\vspace{-0.4cm}
\caption{Pairwise learning framework for DTA prediction.}
\label{PWL}
\end{figure}

\enlargethispage{\baselineskip}
Pairwise learning tasks --- such as drug-target affinity (DTA) prediction \cite{pahikkala2,viljanen}, recommender systems \cite{HerKufTuv2021}, link prediction  \cite{wang2022pairwise}, and information retrieval \cite{Liu2011} --- require models that can efficiently exploit relationships between pairs of objects. 
We call these objects "drugs" ($\drugs \in \mathcal{D}$) and "targets" ($\targets \in \mathcal{T}$), which stems with DTA prediction, but the framework applies to any domain where predictions are made over pairs, although different fields often require distinct evaluation protocols (see, e.g., \cite{pahikkala2,viljanen}). Figure \ref{PWL} shows the pairwise learning framework for DTA prediction.

Currently, neural networks and matrix factorization machines are the two dominant approaches for pairwise learning, particularly in recommender systems, where the goal is to predict labels in the "green area" of Figure \ref{PWL} \cite{xu2021rethinking}. Despite their widespread use, their effectiveness in {\em zero-shot learning} --- corresponding to the "red rectangle" in Figure \ref{PWL}, where both objects in a pair are unseen during training --- remains limited and not thoroughly investigated. Kernel methods have emerged as a strong alternative in these challenging regimes. They are particularly competitive in cold-start scenarios (the "yellow and orange bars" in Figure \ref{PWL}) and zero-shot learning \cite{AIpaper2025}. Recent studies further suggest that kernel methods can match or even surpass deep neural networks on small datasets \cite{abedsoltan2023largekernelmodels}. Moreover, large kernel models that decouple model size from dataset size have been proposed to enable training on millions of samples, showing that kernel methods can scale to levels comparable with modern deep learning systems \cite{abedsoltan2023largekernelmodels,Meanti2020,falkon2017}.

When both drugs and targets have their own feature representations or kernels --- referred to as {\em side information} and illustrated by the "purple areas" in Figure \ref{PWL} --- pairwise kernels provide a natural way to model their affinities. The simplest example is the Kronecker product kernel, which explicitly represents all pairwise affinities between features \cite{viljanen}. Constructing this kernel matrix explicitly is computationally infeasible, as its size grows quadratically with the number of pairs. However, efficient matrix-vector multiplication can be achieved using the {\em generalized vec trick} (GVT) \cite{airola,viljanen}, a sparse Kronecker product algorithm that reduces the cost of such operations from quadratic ($\mathcal{O}(n^2)$, where $n$ is the number of observed pairs) to a form that scales essentially linearly with the number of drugs and targets ($\mathcal{O}(nm+nq)$, where $m$ and $q$ are the numbers of unique drugs and targets, respectively).
This approach allows pairwise kernel methods to scale to much larger datasets than was previously possible, since typically $m,q \ll n$. Nevertheless, the computation of large pairwise kernels can still become a bottleneck, and existing approaches to large kernel models \cite{abedsoltan2023largekernelmodels,abedsoltan2024fasttraininglargekernel,dai2015scalablekernelmethodsdoubly,9180084,Meanti2020,falkon2017} primarily target single-instance kernels without leveraging the structural properties of pairwise kernels.

%Kernel methods have long been a natural choice for these problems, providing both theoretical guarantees and strong empirical performance \cite{viljanen}. However, the key limitation has been scalability: the pairwise kernel matrix grows quadratically in the number of pairs, making traditional approaches computationally prohibitive for large datasets.

In this work, we introduce the {\em stochastic generalized vec trick} (sGVT), a stochastic extension of the GVT for efficient multiplication of Kronecker-product kernel matrices with vectors. Building on sGVT, we further develop \spklns, a scalable pairwise kernel learning algorithm that combines sGVT with a new stochastic optimizer, the {\em stochastic inexact limited-memory bundle method} (\slmbmns), which is based on the inexact nonsmooth optimization solver InexactLMBM \cite{InexactLMBM2026}. %\footnote{We note that \slmbm introduced here should not be confused with the SLMBA introduced in \cite{bigclust2025}; the two algorithms are entirely different, although both are based on the nonsmooth optimization solver LMBM \cite{HaaMieMak:2004,HaaMieMak:2007}.}.
This integration allows \spkl to train kernel-based pairwise models using mini-batches of pairs, thereby reducing both computational and memory costs while preserving the expressive power of pairwise kernels. 
%We provide a {\color{red} theoretical analysis} of sGVT and \slmbmns, and demonstrate through experiments on real-world datasets that \spkl achieves accuracy competitive with state-of-the-art methods, while scaling kernel-based pairwise learning to problem sizes that were previously infeasible.
We analyze the proposed \sgvt and \slmbm methods, and demonstrate through experiments on real-world datasets that \spkl achieves accuracy competitive with state-of-the-art approaches while scaling kernel-based pairwise learning to problem sizes that were previously infeasible. An open-source implementation of \spkl (including \sgvt and \slmbmns) is given at \url{https://github.com/napsu/SPaiK}.

\section{Preliminaries for Pairwise Kernel Learning}
\label{sec:bgpkl}
In this section, we provide the theoretical background on pairwise learning problems and kernel-based methods. In addition, we recall GVT \cite{airola,viljanen}.

\subsection{Pairwise Learning Setup}
In conventional supervised learning, the task is to infer an unknown function $f:\mathcal{X} \rightarrow \mathcal{Y}$ from a set of training samples $\{(x_i, y_i)\}_{i=1}^n$, where each $x_i \in \mathcal{X}$ is an input and $y_i \in \mathcal{Y}$ is its corresponding output. The aim is to construct a prediction function $f$ that generalizes well to unseen inputs $x \in \mathcal{X}$. For regression tasks, the output space is typically $\mathcal{Y} = \mathbb{R}$, whereas for binary classification it is $\mathcal{Y} = \{0,1\}$. 
%In what follows, we consider almost solely  the regression problems.
%
In pairwise learning (see, Figure \ref{PWL}), the training data consist of pairs of 
objects $\x=(\drugs,\targets)$ and their associated labels $\y$, which may be real-valued or categorical. 

The general pairwise kernel learning setting is governed by two assumptions, which characterize the nature of the data and the expected form of the prediction function:

\begin{assumption}\label{pda} {\em (Pairwise Data Assumption)} Each object (e.g., drug or target) typically occurs in multiple pairwise samples in the training set.
\end{assumption}
\begin{assumption}\label{na} {\em (Nonlinearity Assumption)}
The prediction function $f$ typically cannot be decomposed into separate functions of the individual components; that is, it generally cannot be written in the form $f(d,t) = f_d(d) + f_t(t)$ for some functions $f_d$ and $f_t$.
\end{assumption}
\vspace{0.2cm}
 
Assumption~\ref{pda} implies, for instance, that the same drug $d_i$ may appear in multiple pairs $(d_i, t_j)$ and $(d_i, t_k)$ with different targets. Assumption~\ref{na} reflects the empirical observation that DTA (and other pairwise problems) often depend on complex interaction effects \cite{AIpaper2025}. While additive models are mathematically possible, they fail to capture these interactions adequately in most real-world settings. For instance, under additivity, if $f(d_i, t_j) > f(d_k, t_j)$ for some pair of drugs $d_i$ and $d_k$, this ordering would have to hold uniformly across all targets $t_l$. In practice, such uniform behavior is rare, as the relative effectiveness of drugs often varies depending on the specific target.

%We formulate the learning problem as a regularized empirical risk minimization (ERM) task:
%\begin{align}
%\label{kaava1}
%f^* = \underset{f \in \mathcal{H}}{\arg\min} \, \mathcal{L}(\preds, \y) + \lambda \mathcal{R}(f),
%\end{align}
%where $\preds \in \R^n$ are the predicted and $\y \in \R^n$ the correct outputs for the training set, $\mathcal{L}$ is a non-negative convex loss function, $\mathcal{R}$ is a regularization term, and $\lambda > 0$ is a regularization parameter. The function class $\mathcal{H}$ is the hypothesis space.

We formulate the learning problem as a regularized empirical risk minimization (ERM) task:
\begin{align}
\label{kaava1}
f^* = \underset{f \in \mathcal{H}}{\arg\min} , \mathcal{L}(\preds, \y) + \lambda \mathcal{R}(f),
\end{align}
where $\preds \in \R^n$ are the predicted outputs and $\y \in \R^n$ the corresponding correct outputs for the training set, $\mathcal{L}$ is a non-negative convex loss function, $\mathcal{R}$ is a regularization term, and $\lambda > 0$ is a regularization parameter. The function class $\mathcal{H}$ is the hypothesis space.

A standard choice for problem \eqref{kaava1} is ridge regression (regularized least squares, RLS), with the squared error loss $\mathcal{L}(\preds,\y) = \norm{\y-\preds}^2$, Euclidean norm regularization $\lambda \mathcal{R}(f) = \frac{\lambda}{2} \norm{f}^2_\mathcal{H}$, and the reproducing kernel Hilbert space (RKHS) as the hypothesis space $\mathcal{H}$ \cite{PoggioSmale2003}. 
In this work, however, we use the $\varepsilon$-insensitive squared loss \vspace{-2mm}
\begin{align}
\label{eq_eps_insensitive}
\mathcal{L}(\preds,\y)=\frac{1}{2n}\sum\limits_{i=1}^n\max \left(0,(p_i-y_i)^2-\varepsilon\right),
\end{align}
where $\varepsilon > 0$, together with $\ell_1$ regularization. This choice is motivated by both practical and empirical considerations. Since our underlying optimization solver, \slmbmns, is designed for nonsmooth optimization (see, e.g., \cite{BagKarMak:2014}), the nonsmoothness of the loss and regularizer is not a limitation. Moreover, previous experiments on sparse pairwise kernel learning \cite{Kar_pwl_2025} showed that models trained using the $\varepsilon$-insensitive squared loss achieved slightly higher accuracy and faster convergence than those trained with the standard squared loss. In addition, although sparsity is not the primary goal here, the $\ell_1$-norm tends to produce sparser solutions, which has been observed \cite{Kar_pwl_2025} to improve performance in challenging cold-start and zero-shot settings.

%The use of the $\ell_1$ regularizer and the $\varepsilon$-insensitive squared loss instead of the standard RLS formulation is motivated by both practical and empirical considerations. Since our underlying optimization solver, \slmbmns, is designed for nonsmooth optimization, the nonsmoothness of the loss and regularizer is not a limitation. Moreover, previous experiments on sparse pairwise kernel learning \cite{Kar_pwl_2025} showed that models trained using the $\varepsilon$-insensitive squared loss achieved slightly higher accuracy and faster convergence than those trained with the standard squared loss. In addition, although sparsity is not the primary goal here, the $\ell_1$-norm tends to produce sparser solutions, which has been observed \cite{Kar_pwl_2025} to improve performance in challenging cold-start and zero-shot settings.

\subsection{Kernel-Based Formulation}
Kernel methods offer a flexible framework for learning nonlinear functions by mapping input data into high-dimensional feature spaces \cite{airola,hofmann}. When each component of a pair --- such as drug and target --- has its own feature representation or kernel function, we can define a pairwise kernel that operates on the product space.
A straightforward construction is to concatenate the individual feature vectors and apply a standard kernel (e.g., polynomial kernel). However, the Kronecker product kernel is the simplest kernel that models actual pairwise interactions between drug and target features \cite{viljanen}. This kernel enables learning models that generalize to new pairs involving entirely unseen objects --- 
a scenario known as zero-shot learning \cite{airola} (the "red rectangle" in Figure \ref{PWL}).

%A straightforward construction is to concatenate the individual feature vectors and apply a standard kernel (e.g., polynomial kernel). However, to capture the interaction between components more effectively, the Kronecker product kernel is often used \cite{viljanen}\footnote{The Kronecker product kernel is the simplest kernel that models actual pairwise interactions between drug and target features \cite{viljanen}.}. This kernel enables learning models that generalize to new pairs involving entirely unseen objects --- 
%a scenario known as zero-shot learning \cite{airola} (the “red rectangle” in Figure \ref{PWL}).

Let $k_\mathcal{D}: \mathcal{D} \times \mathcal{D} \rightarrow \mathbb{R}$ and $k_\mathcal{T}: \mathcal{T} \times \mathcal{T} \rightarrow \mathbb{R}$ be positive semidefinite kernels for drugs and targets, respectively. 
%\footnote{In this work, the kernel functions $k_\mathcal{D}$ for drugs and $k_\mathcal{T}$ for targets come from the chemical structure and sequence similarity matrices, respectively.}
The Kronecker product kernel $k_{\mathcal{D},\mathcal{T}}:(\mathcal{D} \times \mathcal{T})\times(\mathcal{D} \times \mathcal{T})\rightarrow\R$ on pairs $(d, t) \in \mathcal{D} \times \mathcal{T}$ is defined as
\[
k_{\mathcal{D},\mathcal{T}}((d,t),(d',t')) = k_\mathcal{D}(d,d') \cdot k_\mathcal{T}(t,t').
\]
In addition, let $K \in \mathbb{R}^{n \times n}$ denote the pairwise Kroneker product kernel matrix for all $n$ training pairs, with entries
\[
K_{i,j} = k_{\mathcal{D},\mathcal{T}}((d_i, t_i), (d_j, t_j)) \qquad i,j =1,\ldots,n.
\]
Then, under the representer theorem \cite{scholkopf2}, the minimizer $f^*$ of \eqref{kaava1} admits the representation
\[
f(d, t) = \sum_{i=1}^n a_i \, k_{\mathcal{D},\mathcal{T}}((d_i, t_i), (d, t)),
\]
where $\duals \in \mathbb{R}^n$ are dual coefficients. Additionally, predictions for the training data can be compactly written as $\preds = K \duals$  \cite{viljanen}.
% Pitäiskö mainita  RKHS associated with $k_{\mathcal{D},\mathcal{T}}$ as the hypothesis space?

Assumption \ref{pda} implies that $n \gg m + q$ with $m$ and $q$ denoting the number of unique drugs and targets, respectively. This fact is used to develop computational short-cuts tailored specifically to pairwise kernel learning problems in \cite{airola,viljanen}. 
  More precisely, the large kernel matrix $K$ needs not be computed explicitly 
%($\mathcal{O}(n^2)$) 
but the matrix-vector product $K\duals$
can be computed implicitly in  $\mathcal{O}(nm+nq)$ time using a GVT --- a sparse Kronecker product multiplication algorithm --- introduced in \cite{airola}.
  
 \subsection{Generalized Vec Trick}
We start this section by giving a computational shortcut called Vec Tric (Roth's column lemma \cite{Roth1934}). Let $M \in \R^{a \times b}$, $Q \in \R^{b \times c}$, and $N \in \R^{c \times d}$. Then 
\begin{align}\label{vt}
(N^\top \otimes M) {\rm vec}(Q) = {\rm vec}(MQN),
\end{align}
where ${\rm vec}(A) \in \R^{ab \times 1}$ is the vectorization of $A \in \R^{a \times b}$ obtained by stacking all the columns of A in order starting form the first. In practice, computing the vectorized form in \eqref{vt} is much more efficient, as it avoids the direct computation of the large Kronecker product. 

In \cite{airola}, this idea is further accelerated by exploiting the sparsity of the training labels.
Let $R(\drugs,\targets)\in \R^{n \times \mathcal{D}\times \mathcal{T}}$ denote the Kronecker product index matrix given by
\begin{align}\label{KroneckerIndex}
    R(\drugs,\targets)_{i,(d,t)}=\begin{cases} 1 \quad \text{if }(d,t)=(d_i,t_i)\\
    0 \quad \text{otherwise,}\end{cases}
\end{align}
where, as before, $\drugs \in \ds^n$ and $\targets \in \ts^n$ are sequences of drugs and targets. The drug and target kernels are denoted by 
$D \in \R^{\barm\times m}$ and $T \in \R^{\barq\times q}$,
where $m$ and $q$ represent the numbers of distinct drugs and targets in the training set, and $\barm$ and $\barq$
 are the corresponding numbers in another sample of data $(\bar\drugs, \bar{\targets})$, which may or may not overlap with the training set --- for example, a validation or test set.

Now predictions for another sample $(\bar\drugs, \bar{\targets})$ of size $\barn$ can be computed as \enlargethispage{0.9\baselineskip}
$$\preds \leftarrow R(\bar\drugs, \bar{\targets})(T \otimes D)R(\drugs,\targets)^\top \duals.$$
Using the GVT, this product can be computed in $\mathcal{O}(\min (\bar{q}n +m\bar{n},\bar{m}n +q\bar{n}))$  time \cite{airola,viljanen}.

Predictions for the training data correspond to the special case
$(\bar\drugs, \bar{\targets}) = (\drugs,\targets)$. In this case, they are given as
$$\preds \leftarrow R(\drugs,\targets)(T \otimes D)R(\drugs,\targets)^\top \duals.$$
We recall the GVT algorithm for training in Algorithm~\ref{alg_GVT2}. The algorithm performs training in $\mathcal{O}(nm+nq)$ time. %\cite{airola}.
In what follows, we use the square bracket notation $[a]$ to denote the index set $\{1,\ldots,a\}$ ($a\in \N$), while
$r = (r_1,\ldots,r_n)^\top \in [q]^n$ and $s = (s_1,\ldots,s_n)^\top  \in [m]^n$ denote the sequence of the row indices  of $T$ and $D$ associated to Kronecker product index matrix $R(\drugs,\targets)$. Example \ref{example_rowind} in Section \ref{sec_sGVT} provides further clarification of the index sets. %\vspace{-1mm}
\medskip \enlargethispage{\baselineskip}
%
%$s=(s_1,\ldots,s_f) \in [a]^f$ to denote a sequence of $f$ row indices for matrix $M \in \R^{a \times b}$.

\begin{algorithm}[ht]
{
\caption{GVT for Training \cite{airola}}
\label{alg_GVT2}
{\small
\begin{minipage}{0.95\textwidth}
\begin{description}
\item[Input:] $T\in\R^{q \times q}$, $D\in\R^{m \times m}$, $\duals\in \R^n$, $\br \in [q]^n$, and $\bs \in [m]^n$.
\item[Output:] $\preds = R(T \otimes D)R^\top \duals \in \R^{n}$.\medskip

\item[1.] $M \leftarrow \bzero \in \R^{m \times q}$
\item[2.] {\bf For }$h=1,\ldots,n$ {\bf do}
    
\item[3.]\qquad    $i,j \leftarrow r_h,s_h$
\item[4.]\qquad{\bf For } $k=1,\ldots,q$ {\bf do}
\item[5.]\qquad\qquad $M_{j,k} \leftarrow M_{j,k} + a_h T_{k,i}$ \hfill $\triangleright \,\, \mathcal{O}(nq)$ time operation

\item[6.]  $\preds \leftarrow \bzero \in \R^n$
\item[7.] {\bf For }$h=1,\ldots,n$ {\bf do}
    
\item[8.]\qquad    $i,j \leftarrow r_h,s_h$
\item[9.]\qquad {\bf For} $k=1,\ldots,m$ {\bf do}
\item[10.]\qquad\qquad $p_h \leftarrow p_h + D_{j,k} M_{k,i}$ \hfill $\triangleright \,\, \mathcal{O}(nm)$ time operation
 \item[11.] {\bf Return} $\preds$
\end{description}
\end{minipage}

}
}
\end{algorithm}

\section{Stochastic Generalized Vec Trick}\label{sec_sGVT}
GVT can accelerate any optimization method whose computational cost is dominated by multiplications of a pairwise kernel matrix with a vector. However, the training time of the GVT still directly depends on the number of training samples $n$. In this section, we introduce a stochastic version, sGVT (Algorithm~3), which can carry out kernel matrix multiplication in $\mathcal{O}(n_\mathcal{B} m + n_\mathcal{B} q)$ time, where $n_\mathcal{B}$ is the user specified batch size.

%To model affinities between drug–target pairs, the batch in \sgvt must be selected either target- or drug-wise --- that is, the training data are considered block-wise. Specifically, in the target-wise selection, we randomly choose $q_\batch \leq q$ target indices, and include the drug indices that interact with these targets. This yields a batch $\batch$ containing $n_\batch$ indices. An illustration of the target-wise batch selection procedure is provided in Example~\ref{example_batch}.

To model affinities between drug-target pairs, the batch in \sgvt must be selected either target- or drug-wise --- that is, the training data are considered block-wise. In what follows, we consider only the target-wise batch selection. 
Specifically, we randomly choose $q_\batch \leq q$ target indices  and include the drug indices  that interact with these targets. This yields a batch $\batch$ containing $n_\batch$ indices. An illustration of the target-wise batch selection procedure is provided in Example~\ref{example_batch}.
%Specifically, we randomly choose $q_\batch \leq q$ target indices ($m_\batch \leq m$ drug indices), and include the drug indices (target indices) that interact with these targets (drugs). This yields a batch $\batch$ containing $n_\batch$ indices. An illustration of the target-wise batch selection procedure is provided in Example~\ref{example_batch}.

\begin{figure}[ht]
\begin{center}
\includegraphics[width=0.390\linewidth]{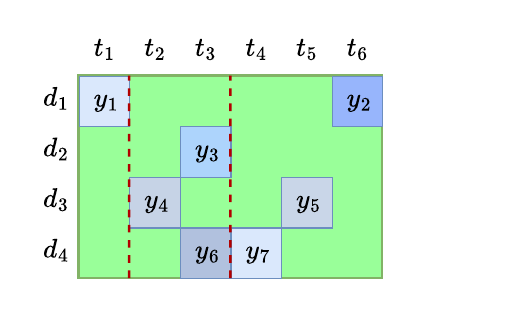}    
\end{center}
\vspace{-0.4cm}
\caption{Training data for Examples \ref{example_batch} and \ref{example_rowind}.}
\label{fig_example}
\end{figure}

\begin{example}\label{example_batch} 
\textbf{(Target-wise batch selection)}.
Assume the training data contains $m=4$ drugs $\{d_1,\dots,d_4\}$ and $q=6$ targets $\{t_1,\dots,t_6\}$ with $n=7$ observed drug-target affinities $\{y_1,\dots,y_7\}$ (see Figure \ref{fig_example}).  
Suppose we randomly select a batch of $q_\batch=2$ targets, say $\{t_2,t_3\}$.  
The batch $\mathcal{B}$ is then defined as the set of indices $i$ such that the corresponding affinity $y_i$ involves a target in $\{t_2,t_3\}$.  
 
For instance, in Figure \ref{fig_example} the known affinities include  
\[
\{(d_1,t_1),\,(d_1,t_6),\,(d_2,t_3),\,(d_3,t_2),\,(d_3,t_5),\,(d_4,t_3),\,(d_4,t_4)\}
\]   
with indices $1,\dots,7$, respectively. Thus, the resulting batch is
 $\mathcal{B} = \{3,4,6\}$, which corresponds to pairs
\[
 \{(d_2,t_3),\,(d_3,t_2),\,(d_4,t_3)\},
\]  
and  the batch size is $n_\mathcal{B}=3$, determined by the number of selected drug-target pairs matching the chosen targets.  
\end{example}

\begin{remark}
While one could also construct batches in a drug-wise manner, our preliminary experiments suggest that the target-wise strategy consistently performs better, producing more accurate predictions in pairwise learning tasks --- particularly in DTA prediction. We note, however, that this effect may be problem specific. 
\end{remark}

In addition to batch selection, another key component of \sgvt is the maintenance of the {\em auxiliary dual-drug matrix} $M \in \R^{m \times q}$. %In the case of drug-wise batch selection, this matrix should instead be the auxiliary dual–target matrix and the order of computations in Algorithm~\ref{alg_SGVT2} should be adjusted accordingly. 
As noted earlier, predictions for the training data can be expressed compactly as $p = Ka$. In \sgvtns, however, we compute and update $p_i$ only for indices $i$ in the current batch $\batch$, while for all other indices $p_j$ we reuse their values from previous iterations. Nevertheless, each $p_i$ ($i \in \batch$) may depend on all dual variables $a_j$ ($j=1,\ldots,n$). 
%The role of matrix $M$ is to capture these dependencies between predictions in the current batch and dual variables outside it.
The aim of matrix $M$ is to model dependencies between predictions in the current batch and dual variables outside the current batch. Preliminary experiments indicate that this information-preserving mechanism is essential for the performance of \sgvtns: without the matrix $M$, the stochastic approximation loses important interaction information, leading to a substantial deterioration in predictive accuracy.

The last ingredient needed before introducing the \sgvt algorithm is the definition of index matrices for the kernel matrices. In line with Algorithm~\ref{alg_GVT2}, we use the sequences $\bs = (s_1,\ldots,s_n)^\top \in [m]^n$ and $\br = (r_1,\ldots,r_n)^\top \in [q]^n$, but, as with the predictions, we restrict attention to indices belonging to the current batch. Example~\ref{example_rowind} illustrates the selection of these indices.

\begin{example}\label{example_rowind} 
\textbf{(Kronecker product index matrix and row indices of $D$ and $T$)}.
Assume again that the training data contains of $m=4$ drugs $\{d_1,\dots,d_4\}$ and $q=6$ targets $\{t_1,\dots,t_6\}$, with $n=7$ observed drug-target affinities  $\{y_1,\dots,y_7\}$ (see Figure \ref{fig_example}).
Then sequences $\bs = (s_1,\ldots,s_n)^\top \in [m]^n$ and  $\br =(r_1,\ldots,r_n)^\top \in [q]^n$ specify the row indices of $D$ and $T$, respectively, and are given by   
$$
s = (1,1,2,3,3,4,4)^\top \qquad \text{and} \qquad r = (1,6,3,2,5,3,4)^\top.
$$
The index matrix $R(d,t)$ (see, eq.\ \eqref{KroneckerIndex}) for the Kronecker product $T \otimes D$ can then be written as 
$$
R(d,t) = \begin{pmatrix}
    e^\top_{(r_1 - 1)m + s_1 }\\
    \vdots \\
    e^\top_{(r_n - 1)m + s_n }
\end{pmatrix} = \begin{pmatrix}
    e^\top_{1} &
    e^\top_{21} &
    e^\top_{10} &
    e^\top_{7}&
    e^\top_{19}&
    e^\top_{12}&
    e^\top_{16}
\end{pmatrix}^\top,
$$
where $e_i$ denotes the $i$th standard basis vector in $\R^{mq}$.

Further, consider the batch $\batch = \{3,4,6\}$ defined in Example~\ref{example_batch}. The corresponding row-index sequences are
$$
s_\batch = (2,3,4)^\top \qquad \text{and} \qquad r_\batch = (3,2,3)^\top,
$$
which specify the rows of $D$ and $T$ relevant for constructing the Kronecker product index matrix. However, we do not explicitly form a separate index matrix for the batch.

%Suppose we randomly select a batch of $q_\batch=2$ targets, say $\{t_2,t_3\}$.  
%The batch $\mathcal{B}$ is then defined as the set of indices $i$ such that the corresponding interaction $y_i$ involves a target in $\{t_2,t_3\}$.  
 
%For instance, in Figure \ref{fig_example} the known interactions include  
%\[
%\{(d_1,t_1),\,(d_1,t_6),\,(d_2,t_3),\,(d_3,t_2),\,(d_3,t_5),\,(d_4,t_3),\,(d_4,t_4)\}
%\]   
%with indices $1,\dots,7$, respectively. Thus, the resulting batch is
% $\mathcal{B} = \{3,4,6\}$, which corresponds to pairs
%\[
% \{(d_2,t_3),\,(d_3,t_2),\,(d_4,t_3)\},
%\]  
%and  the batch size is $n_\mathcal{B}=3$, determined by the number of selected drug-target pairs matching the chosen targets.  
\end{example}

Next, we present the sGVT training algorithm, assuming target-wise batch selection. In the case of drug-wise batch selection, the order of computations must be modified accordingly, and the auxiliary dual-drug matrix should be replaced by the auxiliary dual-target matrix.
\begin{algorithm}[ht] 
{
\caption{Stochastic GVT for Training} % Tätä kääntämässä, vaatii tarkistuksen
\label{alg_SGVT2}
{\small
\begin{minipage}{0.95\textwidth}
\begin{description}
\item[Input:] $T\in\R^{q \times q}$, $D\in\R^{m \times m}$, $M \in \R^{m \times q}$ ($M = \bzero$ at the first iteration), $\duals\in \R^n$, $\preds\in \R^n$, $\br \in [q]^n$, $\bs \in [m]^n$, and batch indices $\batch \in [n]^{n_\batch}$. 
\item[Output:] A new approximation of $\preds = R(T \otimes D)R^\top \duals \in \R^{n}$ and an updated matrix $M$.                                                              \medskip
%\item[1.] $M \leftarrow \bzero \in \R^{m \times q}$
%\item[2.] {\bf For }$h=1,\ldots,n$ {\bf do}
\item[1.] {\bf For all }$h\in \batch$ {\bf do}
    
\item[2.]\qquad    $i \leftarrow r_h$
%\item[1.] {\bf For all }$j \in \bs_\batch$ {\bf do}
%\qquad{\color{red} No need to go trough duplicates}
\item[3.]\qquad{\bf For } $k=1,\ldots,m$ {\bf do}
\item[4.]\qquad\qquad $M_{k,i} \leftarrow 0$ \hfill $ \triangleright \,\, \mathcal{O}(n_\batch m)$ time operation
\item[5.] {\bf For all }$h\in \batch$ {\bf do}
    
\item[6.]\qquad    $i,j \leftarrow r_h,s_h$
\item[7.]\qquad{\bf For } $k=1,\ldots,m$ {\bf do}
\item[8.]\qquad\qquad $M_{k,i} \leftarrow M_{k,i} + a_h D_{k,j}$ \hfill $ \triangleright \,\, \mathcal{O}(n_\batch m)$ time operation
  
\item[9.] {\bf For all }$h\in \batch$ {\bf do}
%\item[7.] {\bf For }$h=1,\ldots,n$ {\bf do}
\item[10.]\qquad    $p_h \leftarrow 0$

\item[11.]\qquad    $i,j \leftarrow r_h,s_h$
\item[12.]\qquad {\bf For} $k=1,\ldots,q$ {\bf do}
\item[13.]\qquad\qquad $p_h \leftarrow p_h + M_{j,k}T_{k,i}$ \hfill $ \triangleright \,\, \mathcal{O}(n_\batch q)$ time operation
 \item[14.] {\bf Return} $\preds$, $M$.
\end{description}
\end{minipage}

} 
}
\end{algorithm}

\begin{remark}
In Steps 4 and 8 of Algorithm \ref{alg_SGVT2}, we initialize and update only the elements of matrix $M$ corresponding to the current batch indices, while leaving the other elements unchanged. This strategy preserves information from previous batches and enables the modeling of dependencies between predictions in the current batch and dual variables outside the current batch. After a few iterations of Algorithm \ref{alg_SGVT2}, each $p_i$, $i=1,\ldots,n,$ depends on almost all $a_j$, $j=1,\ldots,n$, some with old values from previous iterations.
\end{remark}

%\newpage
\section{Pairwise Learning Model}
In this work, we consider the $\varepsilon$-insensitive squared loss together with $\ell_1$ regularization\footnote{We note, however, that our approach is not tied to the chosen loss function; any other locally Lipschitz continuous loss function could equally well be applied.}. In the kernel representation, this leads to the minimization problem
\begin{align}
\label{drproblem}
\underset{{\duals\in\R^n}}{\text{minimize}} \quad
J(\duals)=\mathcal{L}(K\duals,\y) + \lambda \lVert \duals \rVert_1,
\end{align}
where $K$ is the pairwise kernel matrix, $\duals \in \R^n$ is the vector of coefficients, and $\lambda > 0$. The loss function $\mathcal{L}(K\duals,\y)$ is defined in \eqref{eq_eps_insensitive} with $\varepsilon > 0$.

Instead of solving \eqref{drproblem} directly with respect to the full dual variable $a \in \R^n$, we repeatedly minimize it over random batch-specific dual variables $a_\batch \in \R^{n_\batch}$. In other words, only the components of $a$ corresponding to the current batch indices are treated as variables, while the remaining components are kept fixed. Consequently, we compute $K a$ only for indices in the current batch, and for all others we reuse the values $p_i$ and $a_i$ from previous iterations. The auxiliary dual-drug matrix $M \in \R^{m \times q}$ (together with the auxiliary gradient-drug matrix $G \in \R^{m \times q}$, see Algorithm~\ref{alg_PKA}) stores information from earlier batches and thereby captures dependencies between predictions in the current batch and dual variables outside it. 
We denote the resulting batch-wise objective by $J_\batch(a)$ and its corresponding subgradient by $\bxi_{\batch} \in \partial \Jobj_{\batch} (\duals)$, using the full dual vector $a$ in the notation to emphasize that its current value depends on all components, even though only a subset is optimized at each step.

The proposed procedure consists of four main components:
\begin{enumerate}
\item \spkl (Algorithm~\ref{alg_PKA}): The main algorithm \spkl performs the necessary initializations, validates the results, and repeatedly calls \slmbm to solve problem~\eqref{drproblem}. It returns the predictions $p \in \R^n$ together with performance index values (see Section~\ref{datasection} for details).
\item \slmbm (Algorithm~\ref{alg_SLMBA}): \slmbm is responsible for selecting (and reselecting) target-wise batches for the method. It then calls \iLMBM to solve the batch subproblem $\min J_\batch(a)$ for the chosen batch indices.  

\item \iLMBM \cite{InexactLMBM2026}: \iLMBM is a nonsmooth optimization solver designed for large-scale problems with inexact information\footnote{\iLMBM belongs to the \lmbm family \cite{HaaMieMak:2004,HaaMieMak:2007}, whose variants have been successfully applied in several machine learning tasks, including clustering and pairwise kernel learning \cite{Haletal:2023,KarBagTah:2018,bigclust2025,Kar_pwl_2025,missing_values2022,pauliina}. Moreover, its ability to handle inexact function and subgradient information make it particularly suitable for the present stochastic setting.}. 
It solves the batch subproblem iteratively, and in each iteration the computation of $J_\batch(a)$ requires calling the \sgvt algorithm.  

\item \sgvt (Algorithm~\ref{alg_SGVT2}): \sgvt efficiently carries out the computations required for the batch objective via stochastic generalized vec trick updates.  
\end{enumerate}

%the auxiliary dual–target matrix M
%the auxiliary gradient–target matrix G

{ % Tämä korjattu 8.6.2026. Vastaa siis inexactLMBMtuloksia
\begin{algorithm}[pht]
\caption{\spkl}
\label{alg_PKA}
{\small
\begin{minipage}{0.95\textwidth}
\begin{description}
%\item[]

\item[Input:\phantom{ii}] Data 
$x = (d,t) \in \mathcal{X}$ with labels $y \in \mathcal{Y}$; divided into training, validation, and test sets. 
\item[Arguments:] \phantom{ii}
Loss function $\mathcal{L}$ (default: epsilon-insensitive squared loss); %\newline  
kernel functions $k_\mathcal{D}$ and $k_\mathcal{T}$ for drugs and targets\footnote{In this work Gaussian kernels for drugs and targets are used but other kernel functions can be applied as well. The pairwise interactions between drug and target features are modelled via the pairwise Kronecker product kernel matrix $K$.}; %\newline  
%loss sensitivity parameter $\varepsilon>0$ (default: $\varepsilon = 10^{-4}$); %\newline  
percentage of targets used in one batch $p_\batch>0$; %\newline   
the maximum number of iterations $it_{\max}^{\rm SPaiK}>0$ (default: $it_{\max}^{\rm SPaiK}=50$); 
{evaluation metrics} $\mathcal{M}=\{\nu_1,\ldots,\nu_k\}$ (default: $\{\text{C-index},\text{ IC-index},\text{ MSE}\}$);
{primary validation metric} $\nu_{\mathrm{val}}\in\mathcal{M}$ for model selection/early stopping (default: C-index); and the maximum number $i_{\mathrm{val}}^{\max} > 0$ of consecutive validation evaluations without improvement (default: $i_{\mathrm{val}}^{\max} = 3$).

%\item[Arguments:] \phantom{ii}
%Loss function $\mathcal{L}$ (default: epsilon-insensitive squared loss); %\newline  
%kernel functions $k_\mathcal{D}$ and $k_\mathcal{T}$ for drugs and targets\footnote{In this work Gaussian kernels for drugs and targets are used but other kernel functions can be applied as well. The pairwise interactions between drug and target features are modelled via the pairwise Kronecker product kernel matrix $K$.}; %\newline  
%loss sensitivity parameter $\varepsilon>0$ (default: $\varepsilon = 10^{-4}$); %\newline  
%percentage of targets used in one batch $p_\batch>0$; %\newline   
%the maximum number of iterations $it_{\max}^{SPaiK}>0$ (default: $it_{\max}^{SPaiK}=10$); 
%{evaluation metrics} $\mathcal{M}=\{\nu_1,\ldots,\nu_k\}$ (default: $\{\text{C-index},\text{ IC-index},\text{ MSE}\}$); and
%{primary validation metric} $\nu_{\mathrm{val}}\in\mathcal{M}$ for model selection/early stopping (default: C-index).

\item[Output:] Predictions $p \in R^n$ and all performance metrics $\nu_i \in \mathcal{M}$ w.r.t.\ test data.
 %the termination limit $i_{term}>1$
 %, the iteration limit $i_{\max}>0$, the number of stored correction vectors $ m_c \geq 3$, and a finite-sum function $\Jobj$ to be minimized.

\item[Step 0.] ({\em Initialization})  Set a starting
  point $\duals^1 \leftarrow 0 \in \R^n$ %. Compute $f(\duals^1) = \mathcal{L}(0,y)$ 
  and compute the regularization parameter $\lambda \leftarrow \frac{\mathcal{L}(0,\y)}{n^2}$ and the loss sensitivity parameter $\varepsilon = 10^{-5} \cdot \max_{y \in \mathcal{Y}_{\rm train}}(y)$.
  Compute the following:
  $$
  q_\batch \leftarrow \max (\lfloor \frac{p_\batch  q}{100}\rfloor,1), \qquad i_{\batch}^{\max} \leftarrow \lfloor\frac{q}{q_\batch}\rfloor, \qquad \text{and} \qquad
  it_{\max}^{\rm StoILMBM} \leftarrow \lfloor\frac{1000}{i_{\batch}^{\max}}\rfloor,
  $$ where $q_\batch$ is the target-wise batch size, 
  $it_{\max}^{\rm StoILMBM} $ is the maximum number of iterations used with \slmbmns, and 
  $i_{\batch}^{\max}$ is the maximum number of batches after which predictions are (re)validated.
 %Compute the target-wise batch size $ q_\batch \leftarrow \frac{p_\batch  q}{100}$, the maximum number of batches $i_{\batch}^{\max} \leftarrow \max(\frac{q}{q_\batch})$, and 
  %the maximum number of iterations $it_{\max}^{\slmbm} \leftarrow \frac{1000}{i_{\batch}^{\max} }$ used with the \slmbm after which the predictions are (re)validated.
  % pitää varmaan käyttää jotain floor merkintää
  % koodissa $it_{\max}=nth = 1000/b_max = 1000/(q/q_\batch)$  ja joitain rajoitteita ettei liian pieni ja b_max = q/q_\batch alussa ja seuraavilla iteraatioilla 1.

  %Compute $p^1=Ka^1$ via GVT and 
    Set $M^1,\, G^1 \leftarrow 0 \in \R^{m \times q}$, $\nu_{\rm best} \leftarrow0$, $\duals_{\rm best}\leftarrow0 \in R^n$, $i_{\mathrm{val}} \leftarrow 0$, and $ i \leftarrow 1$. 
    %and 
    %$i=1$, $t=$ {\em false}, $CI_{best} = 0$.
\item[Step 1.] ({\em Regularized risk minimization problem}) Apply \slmbm (Algorithm \ref{alg_SLMBA}) to minimize the regularized risk minimization problem \eqref{drproblem} starting from $a^i$ with the regularization parameter $\lambda$, matrices $M^i$ and $G^i$, and limits $i_{\batch}^{\max}$ and $it_{\max}^{\rm StoILMBM}$. 
Stop Algorithm \ref{alg_SLMBA} after $it_{\max}^{\rm StoILMBM}$ iterations or when a stationary point of the batch problem is found; in the latter case, set \texttt{stop}  $\leftarrow$  \texttt{true}.
Denote the obtained solution by $a^{i+1}$ and the corresponding matrices $M^{i+1}$ and $G^{i+1}$.

\item[Step 2.] ({\em Validation and early stopping})  Compute the validation metric $\nu_{\rm val}$ w.r.t.\ the validation data and, if $\nu_{\rm val}>\nu_{\rm best}$, set $\nu_{\rm best} \leftarrow \nu_{\rm val}$, $a_{\rm best}  \leftarrow a^{i+1}$, and $i_{\rm val} \leftarrow 0$. Otherwise, set $i_{\rm val} \leftarrow i_{\rm val} +1$. If $i_{\rm val} > i_{\rm val}^{\max}$, proceed to Step~4.

\item[Step 3.] ({\em Update parameters}) If $i < it_{\max}^{\rm SPaiK}$, set $i_{\batch}^{\max} \leftarrow 1$, $i \leftarrow i+1$, and go to Step~1.

\item[Step 4.] ({\em Termination}) 
STOP with $a_{best}$ as a best solution. Compute and return the predictions and performance indices $\nu_i \in \mathcal{M}$ w.r.t.\ test data. 
%Otherwise, set $t=$ {\em false}.

  \end{description}
\end{minipage}

}
\end{algorithm}
}
% koodissa $it_{\max}=nth = 1000/b_max = 1000/(q/q_\batch)$  ja joitain rajoitteita ettei liian pieni
\begin{algorithm}[pht]
{
\caption{\slmbm}
\label{alg_SLMBA}
{\small
\begin{minipage}{0.95\textwidth}
\begin{description}%[leftmargin=0.5cm]
\item[Input:] A starting point $\duals^1 \in \R^n$, 
%{\color{red} and a prediction $\preds^1 \in \R^n$},
the saved matrices $M^1,\,G^1 \in \R^{m \times q}$, the target-wise batch size $ q_\batch \leq q$, the maximum number of batches $i_{\batch}^{\max}>0$, and
%the termination limit $i_{term}>1$, 
the iteration limit $it_{\max}>0$.
%, and the number of stored correction vectors $ m_c \geq 3$.
%, and a finite-sum function $\Jobj$ to be minimized.

\item[Output:]  A heuristic minimizer $\duals^h \in \R^n$ of $\Jobj:\R^n \rightarrow \R$ and the updated matrices $M^h,\,G^h \in \R^{m \times q}$.

\item[Step 0.] {\em Initialization:}
Select a batch $\batch_1$ based on $q_\batch$ randomly selected target indices and their corresponding drug indices. Set $n_\batch \leftarrow |\batch_1|$, $i_\batch \leftarrow 1$, and $h \leftarrow 1$.
%  Set $\Jobj_{best} \leftarrow \infty$ and $h \leftarrow 1$. Initialize the correction matrices $S^1$ and $U^1$ as empty matrices.
  
\item[Step 1.] {\em Solving the batch problem:} Starting from the point $\duals^h$ apply \iLMBM  to minimize the batch function $\Jobj_{\batch_h}$:
\begin{itemize}
    \item Use \sgvt (Algorithm \ref{alg_SGVT2}) with matrices $M^{h}$ and $G^{h}$ to compute $\Jobj_{\batch_h}(\duals^h_s)$ and $\bxi^s_{\batch_h} \in \partial \Jobj_{\batch_h} (\duals^h_s)$ (here, $s$ is an iteration index within \iLMBMNS). Save the \sgvt matrices $M^h_s$ and $G^h_s$.
    \item Stop \iLMBM after $it_{\max}$ iterations or with the stationary point of $J_{\batch_h}$. 
    Denote the solution by $\duals^{h+1}$, and the corresponding matrices by $M^{h+1}$ and $G^{h+1}$.
\end{itemize}
Set $h\leftarrow h+1$.

%\item[Step 0.1.] {\color{red} ({\em Stopping criterion})
\item[Step 2.] {\em Stopping criterion:}
Set $i_\batch \leftarrow i_\batch + 1$. If $i_\batch > i_{\batch}^{\max}$ STOP the algorithm with $\duals^h$ as a best solution, and $M^{h}$ and $G^{h}$ the associated matrices.

%\item[Step 0.5.] {\color{red} ({\em Selection of the batch})    
\item[Step 3.] 
{\em Selection of the new batch:}
Randomly select $q_\batch$ target indices such that each target index is included at least once within an epoch. Select a new batch $\batch_h$ based on these target indices 
and their corresponding drug indices. Set $n_\batch \leftarrow |\batch_h|$, and go to Step 1.

%\item[Step 1.] ({\em Serious step initialization.}) 
\end{description}
\end{minipage}
}}
\end{algorithm}

\begin{remark}%\enlargethispage{2\baselineskip}
\label{rem3}
In the first call of \slmbm (Algorithm \ref{alg_SLMBA}), we use one complete epoch over the target indices when selecting the batches. That is, in Step 0 of Algorithm \ref{alg_PKA}, we set
$
i_{\batch}^{\max} \leftarrow q/q_\batch,
$
and in Step 3 of Algorithm \ref{alg_SLMBA}, the batches are selected such that each target index is included at least once during this epoch, with the target indices processed in random order. This is not necessarily required by the method, but it provides a more accurate results in practice as shown in Section \ref{sec_epoch_vs_random}. In the subsequent calls of \slmbm, we use only one randomly selected batch before re-validation, that is, we set
$
i_{\batch}^{\max} \leftarrow 1
$
in Step 3 of Algorithm \ref{alg_PKA}.
\end{remark}

\begin{remark}
We compute the batch function value $J_\batch(a)$ using the full dual variable $a$. From a purely optimization perspective, one could instead pass only the batch-specific dual variable $a_\batch$ to \iLMBM (see Step 1 of Algorithm \ref{alg_SLMBA}). However, we deliberately use the full dual variable for two reasons: First, the complete function value is required for validation and performance monitoring, making it essential to maintain consistency across batches. Second, \iLMBM relies on function values in its step-size selection and related mechanisms; providing only partial values would introduce unnecessary variability and could impair the stability of the optimization. Using the full $a$ therefore yields more reliable convergence behavior and ensures that the optimization process remains directly comparable across iterations.\end{remark}

\section{Numerical Experiments} 
\label{sec:num}
Now we are ready to compare the proposed \spklns-algorithm with some state-of-the-art methods in pairwise kernel learning.
We start this section by describing the datasets, different experimental settings for data, and the performance measures used. Then we continue with implementational details of the tested algorithms and, finally, we give the results of our experiments.

\subsection{Data, Experimental Settings, and Performance Metrics } %
\label{datasection}
The numerical experiments were performed on seven benchmark DTA datasets, summarized in Table~\ref{tab:datasets}. These datasets were selected because (i) pairwise kernel methods were originally introduced for biomedical affinity prediction and are widely benchmarked in this domain, (ii) they span a range of sizes, label types, and sparsity levels, and (iii) they include feature representations for both drugs and targets, which are necessary for Kronecker kernel-based learning.

Each dataset is represented by drug and target feature matrices  
$X_{d} \in \mathbb{R}^{m \times m}$, $X_{t} \in \mathbb{R}^{q \times q}$, together with an interaction affinity matrix $Y \in \mathbb{R}^{m \times q}$. In Davis, GPCR, Ion Channels, and Enzymes, all DTAs are known, while Metz, KiBA, and Merget are sparse with only a subset of DTAs observed. The numbers of drugs ($m$), targets ($q$), known affinities ($n$), and percentages of observed affinities (\% observed) are reported in Table \ref{tab:datasets}, with further dataset details available in the cited references. In addition, links to all data used in the experiments are available
at \url{https://github.com/TurkuML/IC-index-experiments}.

\begin{table}[ht]
    \caption{dataset characteristics.} 
    \centering 
    \resizebox{0.80\textwidth}{!}{
    \begin{tabular}{lccccll} \hline
        dataset & $m$ & $q$ & $n$ & \% observed & Label type & Ref.\\ \hline
        Davis & 68 & 442 & 30 056 & 100 & Continuous & \cite{davis2011comprehensive} \\
        Metz & 1 421 & 156 & 93 356 & 42 & Continuous & \cite{metz2011} \\
        KIBA & 2 111 & 229 & 118 254 & 24& Continuous & \cite{tang2014making} \\
        Merget & 2 967 & 226 & 167 995 & 25& Continuous & \cite{merget2017profiling} \\
        G protein-coupled receptor (GPCR) & 223 & 95 & 21 185 & 100 & Binary &  \cite{yamanishi2008prediction} \\
        Ion Channels & 210 & 204 & 42 840 & 100 & Binary &  \cite{yamanishi2008prediction} \\
        Enzymes & 445 & 664 & 295 480 & 100 & Binary & \cite{yamanishi2008prediction} \\
        \hline
    \end{tabular}}
    \label{tab:datasets}
\end{table}
\medskip

% Tätä tekstiä kannattanee vielä muokata ICI paperin mukaiseksi
% nyt vain S1 -- S4 muutettu
%Following \cite{pahikkala2,AIpaper2025,park,schrynemackers}, pairwise prediction tasks can be categorized into distinct {\em experimental settings}, depending on the assumed overlap between the training data and the new pairs to be predicted.  
%Given training data \( z \subseteq \mathcal{X} \times \mathcal{Y}= \mathcal{D} \times \mathcal{T} \times \mathcal{Y} \), we denote by \( \mathcal{X}_z \) the set of in-training-set (ITS) drug–target pairs and by \( \mathcal{X} \setminus \mathcal{X}_z \) the set of off-training-set (OTS) pairs. Let \( \mathcal{D}_z \subseteq \mathcal{D} \) and \( \mathcal{T}_z \subseteq \mathcal{T} \) denote the subsets of drugs and targets appearing in the training data, respectively.   
%The OTS pairs can be further divided into four disjoint subsets, defined as
%\[
%\begin{aligned}
%\mathcal{X}^{\text{IDIT}}_z &= (\mathcal{D}_z \times \mathcal{T}_z) \setminus \mathcal{X}_z, \\
%\mathcal{X}^{\text{IDOT}}_z &= \mathcal{D}_z \times (\mathcal{T} \setminus \mathcal{T}_z), \\
%\mathcal{X}^{\text{ODIT}}_z &= (\mathcal{D} \setminus \mathcal{D}_z) \times \mathcal{T}_z, \\
%\mathcal{X}^{\text{ODOT}}_z &= (\mathcal{D} \setminus \mathcal{D}_z) \times (\mathcal{T} \setminus \mathcal{T}_z),
%\end{aligned}
%\]
%corresponding to the {\em IDIT}, {\em IDOT}, {\em ODIT}, and {\em ODOT} experimental settings, respectively.

Following \cite{pahikkala2,AIpaper2025,park,schrynemackers}, pairwise prediction tasks can be categorized into distinct {\em experimental settings}, depending on the assumed overlap between the training data and the new pairs to be predicted. 
We refer to pairs $x = (d,t) \in \mathcal{X}$ not included in the training set as \emph{off-training-set} (OTS) pairs. Four such OTS settings are distinguished:
\smallskip

\begin{itemize}
\item $x \in \mathcal{X}^{IDIT}$ if and only if $d \in X_d$ and $t \in X_t$ (green area in Figure~\ref{PWL});
\item ${x} \in \mathcal{X}^{IDOT}$ if and only if ${d} \in X_d$ but ${t} \notin X_t$ (orange bar at the rightmost position of Figure~\ref{PWL});
\item ${x} \in \mathcal{X}^{ODIT}$ if and only if ${d} \notin X_d$ but ${t} \in X_t$ (yellow bar at the bottom of Figure~\ref{PWL});
\item ${x} \in \mathcal{X}^{ODOT}$ if and only if ${d} \notin X_d$ and ${t} \notin X_t$ (red rectangle in Figure~\ref{PWL}).
\end{itemize}
Hereafter, these are referred to as the {\em IDIT}, {\em IDOT}, {\em ODIT}, and {\em ODOT} settings (or data).
Setting IDIT corresponds to missing value imputation, where both the drug and the target are present in the training set. Settings IDOT and ODIT represent multilabel learning tasks, aiming to predict labels for novel targets and novel drugs, respectively. Setting ODOT, known as zero-shot learning, is the most challenging case in which predictions are required for pairs where neither the drug nor the target appears in the training data. 

Each dataset is randomly partitioned five times into separate training, validation, and test sets (in proportions of 1/3, 1/3, and 1/3), respecting the requirements of each experimental setting. We report results as averages over these splits, and the same partitions are used across all learning methods to ensure comparability.
%The C-index is a rank-based performance measure that is defined as the probability that the predictions for two sample points are in the same order as their real labels. 

The results of our numerical experiments are evaluated using the {\em Concordance Index} (C-index) \cite{gonen2}, the {\em Interaction Concordance Index} (IC-index) \cite{AIpaper2025}, the {\em Mean Squared Error} (MSE, {only in Appendix}), and the measured CPU time.
The C-index (higher is better) is a convenient performance metric when the relative ordering of labels is more important than their exact values, whereas MSE (lower is better) directly measures the accuracy of predicted values. For reference, a constant predictor achieves a trivial performance level of 0.5 with respect to the C-index, while a value of 1 corresponds to perfect prediction \cite{AIpaper2025}.

The IC-index (higher is better) is a more recent performance metric designed to assess whether a pairwise learning algorithm captures nonlinear interactions between object pairs or not. In other words, it indicates whether the model goes beyond learning simple main effects and provides genuine modeling of pairwise relationships. Formally, the IC-index corresponds to computing the C-index between $y_i - y_j$ and $f(x_i)-f(x_j)$ over all row pairs. Linear predictors (e.g., drug-constant or target-constant functions) achieve a trivial IC-index value of 0.5, whereas a value of 1 reflects perfect modeling of nonlinear pairwise interactions \cite{AIpaper2025}.

\subsection{Methods}
The following pairwise kernel learning algorithms are used in our experiments:
\medskip

%\begin{itemize}
%\begin{description}
\noindent
{\bf CGKronRLS:}
We apply the state-of-the-art method {\tt CGKronRLS} \cite{airola} as a benchmark. This method has demonstrated strong real-world performance, ranking among the top-performing methods in the IDG-DREAM Drug-Kinase Binding Prediction Challenge \cite{cichonska2021}.

{{\tt CGKronRLS}} uses the conjugate gradient (CG) method to solve the standard RLS formulation of the pairwise kernel learning problem \eqref{kaava1}: it applies the squared loss $\norm{y-p}^2$ as a loss function and the Euclidean norm $\frac{\lambda}{2} \norm{f}^2_\mathcal{H}$ for regularization. 
The regularization parameter $\lambda$ is selected from the grid  $\{2^{-10}, 2^{-5}, 2^{-4}, 2^{-3}, 2^{-2}, 2^{-1}, 2^0, 2^1, 2^3, 2^4, 2^5, 2^{10}\}$, and the tuning of $\lambda$ is performed separately for each dataset and each setting. 
In {\tt CGKronRLS}, the C-index of the validation data is used to select the best solution with an early stopping procedure \cite{Yao2007}.
%In CG the maximum number of iterations is set to 1000. 

% During the CG iteration, the C-index on the validation data is monitored and the iteration is terminated when the validation performance no longer improves, following the early stopping principle of \cite{Yao2007}. The model obtained at the best validation point before termination is then used as the final model.

{\tt CGKronRLS} is implemented in Python/Cython and it belongs to RLScore \cite{pahikkala4} --- the regularized least-squares machine learning algorithms package --- available at \url{https://github.com/aatapa/RLScore}.
%{\color{red} Pitäiskö mainita Python/Cython? Kannattanee mainita ettei CGKronRLS tuota %harvoja ratkaisuja!!! Joko tässä (implisiittisesti sanottu jo) tai myöhemmin. 
\smallskip

\noindent
{\bf KronSVM:} Further, we use {\tt KronSVM} \cite{airola} --- a hinge loss variant of {\tt CGKronRLS} --- as a bencmark for binary labelled datasets. We note that hinge loss is generally considered more effective for classification than squared error loss and, indeed, in its original introduction \cite{airola}, {\tt KronSVM} demonstrated slightly better predictive accuracy than {\tt CGKronRLS} for binary-labeled data. 
{\tt KronSVM} uses the truncated Newton method to solve the SVM formulation of the classification problem. The parameters for {\tt KronSVM} are selected similarly to {\tt CGKronRLS}.
In addition, {\tt KronSVM} is implemented in Python/Cython, belongs to RLScore\footnote{We note that a newer implementation of {\tt KronSVM} is available in {\tt RLScore}. The experiments reported in this paper were carried out using the earlier version available at the time of our numerical tests.}, and is available at \url{https://github.com/aatapa/RLScore}.
\medskip

\noindent
\textbf{SPaiK:}  
{\tt SPaiK} is an implementation of Algorithm~\ref{alg_PKA}. The percentage of targets used to form a batch is denoted by $p_\batch \in {100, 20, 10, 5, 1}$ and indicated in the method name {\tt SPaiK}$p_\batch$. In particular, {\tt SPaiK}100 corresponds to using the non-stochastic GVT, that is, the full batch.

Unless otherwise stated, {\tt SPaiK}$p_\batch$ uses the epoch-wise batch selection described in Algorithm~\ref{alg_PKA} and Remark \ref{rem3}: target indices are processed in random order so that each target index is included at least once within an epoch. To assess the effect of this batch-selection rule, we additionally consider a fully random target-wise variant with $p_\batch=10$, denoted by {\tt SPaiK-R}10. In this variant, each batch is formed by randomly selecting $q_\batch$ target indices independently of previous batches.

The best result in each run is selected based on the C-index computed on the validation data. For the epoch-wise variants {\tt SPaiK}$p_\batch$, validation is first performed after one full epoch and then after each subsequently selected batch. For the fully random variant {\tt SPaiK-R}10, where no epoch structure is imposed, validation is performed after each randomly selected batch. To account for randomness, the training process is repeated five times with different random seeds for {\tt SPaiK}$p_\batch$ with $p_\batch \in {20,10,5,1}$. Since the data is also randomly split five times, this results in a total of 25 independent runs for each percentage of targets, and the final results are obtained by averaging over these runs. The same protocol is used for {\tt SPaiK-R}10. For $p_\batch=100$, there are no random batches and training is deterministic given a split; we therefore run once per split, resulting in $5$ runs in total, and report the average.

%The best result in each run is selected based on the C-index computed on the validation data. Validation is first performed after one full epoch and then after each randomly selected batch. To account for randomness, the training process is repeated five times with different random seeds for {\tt SPaiK}$p_\batch$ with $p_\batch \in {20,10,5,1}$. Since the data is also randomly split five times, this results in a total of 25 independent runs for each percentage of targets, and the final results are obtained by averaging over these runs. The same protocol is used for {\tt SPaiK-R}10. For $p_\batch=100$, there are no random batches and training is deterministic given a split; we therefore run once per split, resulting in $5$ runs in total, and report the average.

{\tt SPaiK} is implemented using a combination of Python and Fortran 95 via the F2PY interface. The source code of {\tt SPaiK} is available at \url{https://github.com/napsu/SPaiK}.

% {\tt SPaiK} is an implementation of Algorithm~\ref{alg_PKA}. The percentage of targets used to form a batch is denoted by $p_\batch \in \{100, 20, 10, 5, 1\}$ and indicated in the method name {\tt SPaiK}$p_\batch$. In particular, we note that {\tt SPaiK}100 corresponds to using the non-stochastic GVT (full batch). 
%The best result in each case is selected based on the C-index computed on the validation data. Validation is first performed after one full epoch and then after each randomly selected batch. To account for randomness, the training process is repeated five times with different random seeds for {\tt SPaiK}$p_\batch$ with $p_\batch \in \{20,10,5,1\}$. Since the data is also randomly split five times, this results in a total of 25 independent runs with each percentage of targets, and the final results are obtained by averaging over these runs.
%For $p_\batch=100$, there are no random batches and training is deterministic given a split; we therefore run once per split ($5$ runs total) and report the average.
%
%{\tt SPaiK} is implemented in combination of Python and Fortran 95 via the F2PY interface.
%The source code of {\tt SPaiK} is available at \url{https://github.com/napsu/SPaiK}.

\bigskip
All the learning algorithms use Kronecker product kernel matrices to compute the predictions $\preds = K\duals$. With {\tt CGKronRLS} and {\tt KronSVM} the Kronecker product matrices are computed implicitly using the GVT \cite{pahikkala2,viljanen} from RLScore. The proposed \spkl -algorithm employs \sgvt (Algorithm \ref{alg_SGVT2}), introduced in this paper, for training, while the original GVT %(Algorithm \ref{alg_GVT}) 
is used for validation and final predictions on the test set\footnote{For the test data, we use the original GVT to ensure results that are directly comparable with those obtained by other methods. For validation, the choice was made primarily for ease of implementation. In principle, sGVT could be extended without difficulty to both validation and test phases.}.
The drug and target kernels are computed via Gaussian RBF kernels with the kernel width parameter $\mu=10^5$ as recommended in \cite{airola}. 

Computational experiments are carried out on iMac, 4.0 GHz Intel(R) Core(TM) i7 machine with 16 GB of RAM. We use Python 3.7 and gfortran to compile the Fortran codes.

\subsection{Results}
The results are given in Tables~\ref{tab:davis}--\ref{tab:e} in the appendix and summarized here in Figures \ref{fig1} -- \ref{fig3}.
%We run the proposed \spkaa with four desired levels of sparsity: 0\% ($k=n$), 50\% ($k=n/2$), 80\% ($k=n/5$), and 90\% ($k=n/10$). 
In Figure \ref{fig1} we give the
C-indices (blue bar), IC-index (green bar), %MSE (yellow line), 
and time in seconds (red line) for different algorithms and different batch sizes in continuously labeled datasets (see also Tables \ref{tab:davis}--\ref{tab:merget}).
%The desired sparsity level used with \spkaa is 80\% and it was always reached but once with {\tt DREIS} in Davis data under setting S4. The results obtained with the other desired levels of sparsity are very similar but with the 90\% sparsity, there were somewhat more infeasible solutions (see Tables \ref{tab:davis}--\ref{tab:merget}). Note that {\tt CGKronRLS} only produces dense solutions, while the final sparsity levels obtained with {\tt LMBMKron$\ell_0$LS} are given in the caption of the figure.
In Figure \ref{fig2}, we summarize the corresponding results (see Tables \ref{tab:gpcr}--\ref{tab:e}) for binary labeled data, and in Figure~\ref{fig3}, we compare the epoch-wise and fully random batch selection rules used in {\tt SPaiK}.

\iffalse
%%\begin{landscape}
\begin{figure} [ht]  
% %\centering
  \subfigure[Davis]{
   \includegraphics[width=0.50\textwidth]{{chartDavis}}}%\hspace{-3.7em}
  \subfigure[Metz]{
   \includegraphics[width=0.50\textwidth]{{chartMetz}}}\\%\hspace{-3.7em}
  \subfigure[KiBA]{
   \includegraphics[width=0.50\textwidth]{{chartKiba}}}%\hspace{-3.7em}
  \subfigure[Merget]{\label{fig1d}
   \includegraphics[width=0.50\textwidth]{{chartMerget}}}
\caption{C-index (blue bar), IC-index (green bar), MSE (yellow line), and time (red line) of pairwise kernel learning algorithms in data with continuous labels under different settings.} 
\label{fig1}
\end{figure} 

\begin{figure} [ht]  
% %\centering
  \subfigure[Davis]{
   \includegraphics[width=0.50\textwidth]{{chartDavis2}}}%\hspace{-3.7em}
  \subfigure[Metz]{
   \includegraphics[width=0.50\textwidth]{{chartMetz2}}}\\%\hspace{-3.7em}
  \subfigure[KiBA]{
   \includegraphics[width=0.50\textwidth]{{chartKiba2}}}%\hspace{-3.7em}
  \subfigure[Merget]{\label{fig1d}
   \includegraphics[width=0.50\textwidth]{{chartMerget2}}}
\caption{C-index (blue bar), IC-index (green bar), and time (red line) of pairwise kernel learning algorithms in data with continuous labels under different experimental settings.} 
\label{fig1}
\end{figure} 
\fi

\begin{figure} [ht]  
% %\centering
  \subfigure[Davis]{
   \includegraphics[width=0.50\textwidth]{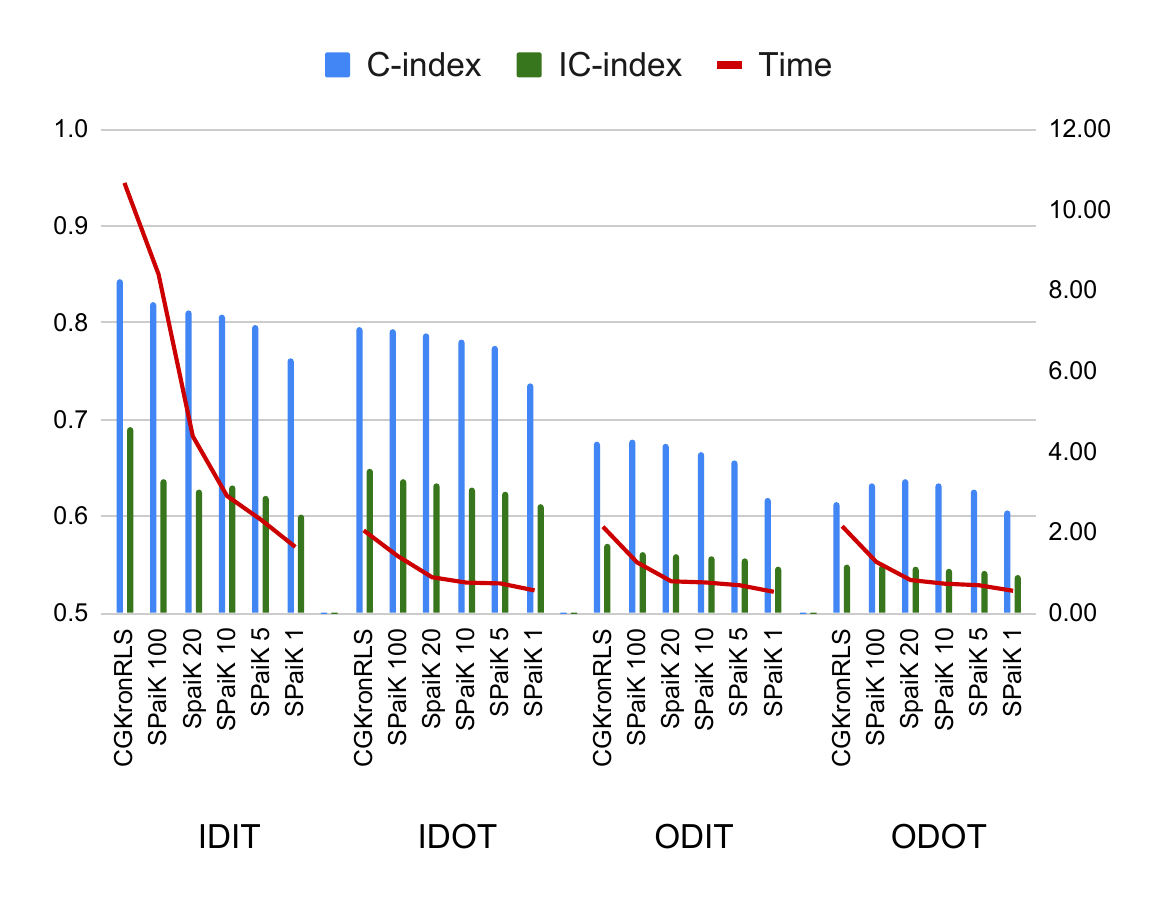}}%\hspace{-3.7em}
  \subfigure[Metz]{
   \includegraphics[width=0.50\textwidth]{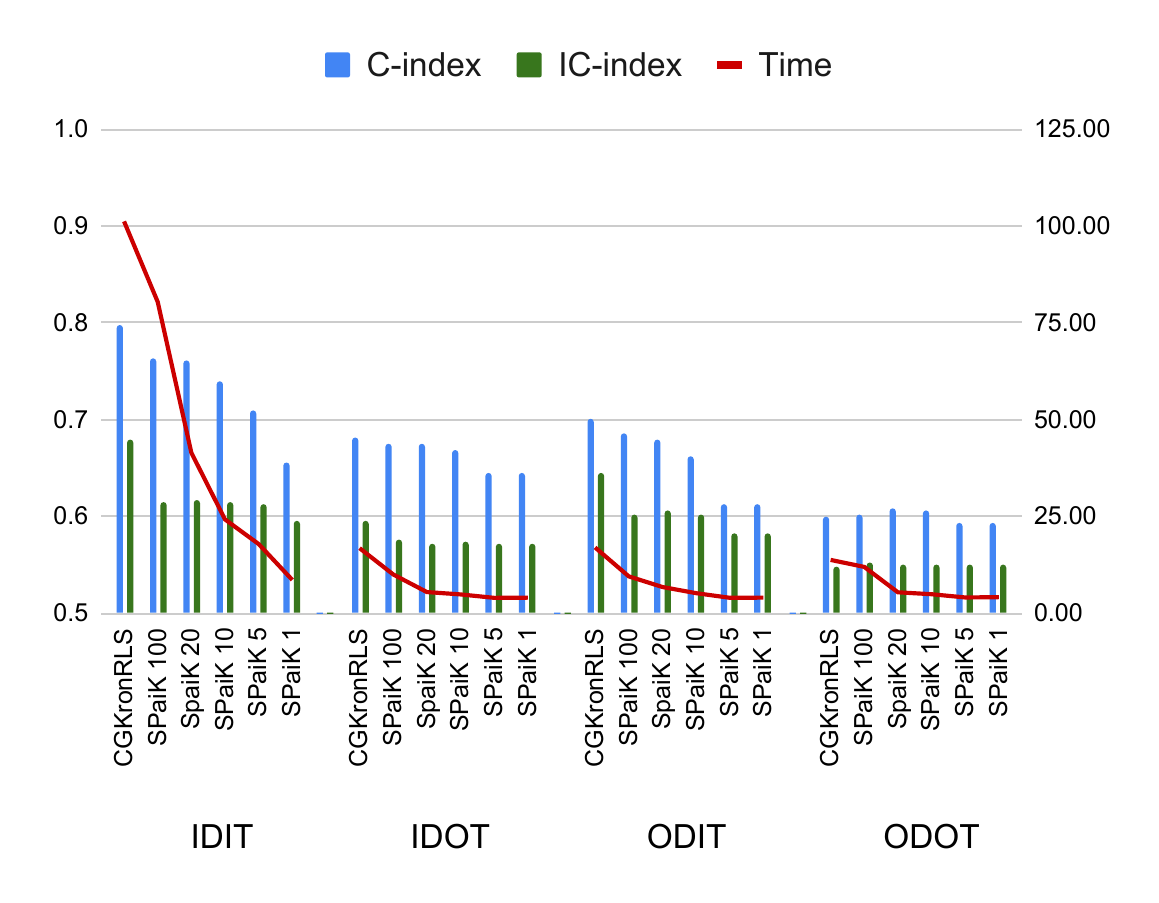}}\\%\hspace{-3.7em}
  \subfigure[KiBA]{
   \includegraphics[width=0.50\textwidth]{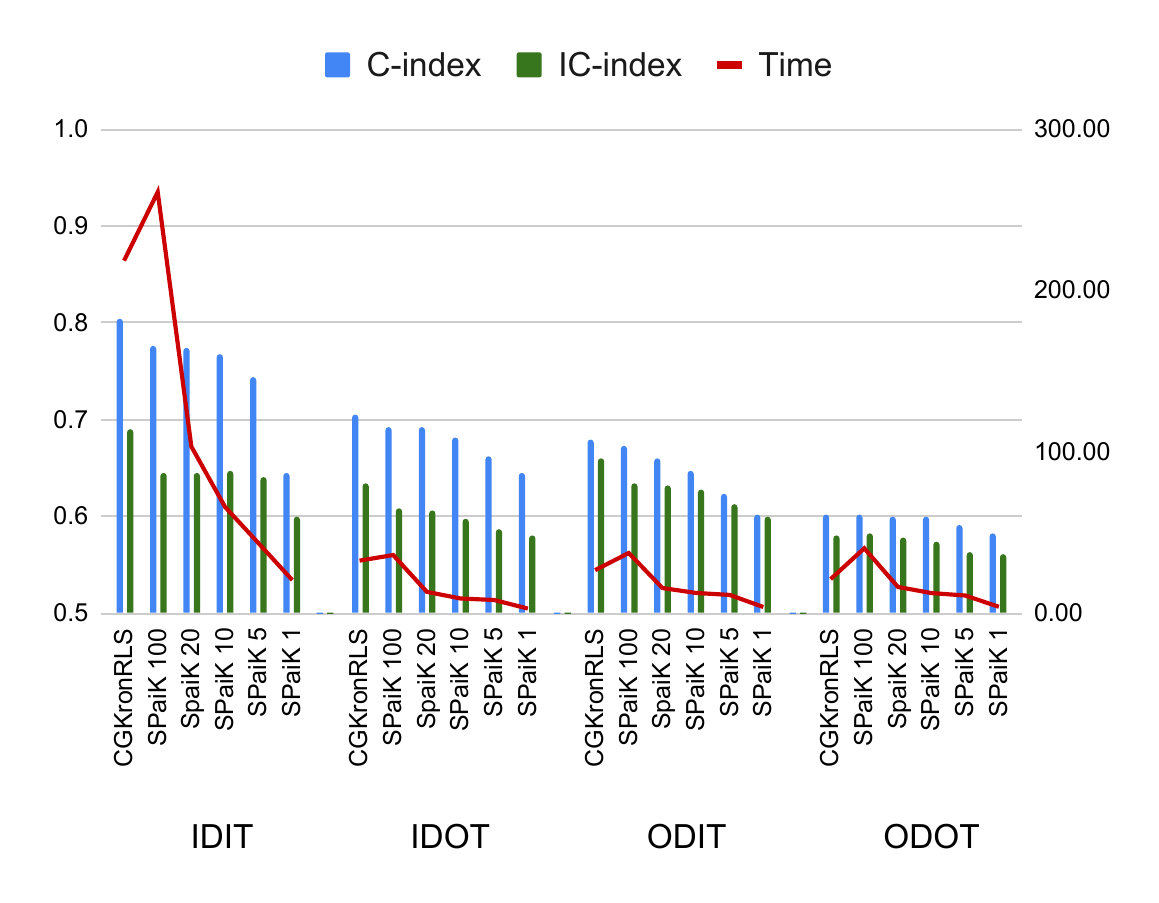}}%\hspace{-3.7em}
  \subfigure[Merget]{\label{fig1d}
   \includegraphics[width=0.50\textwidth]{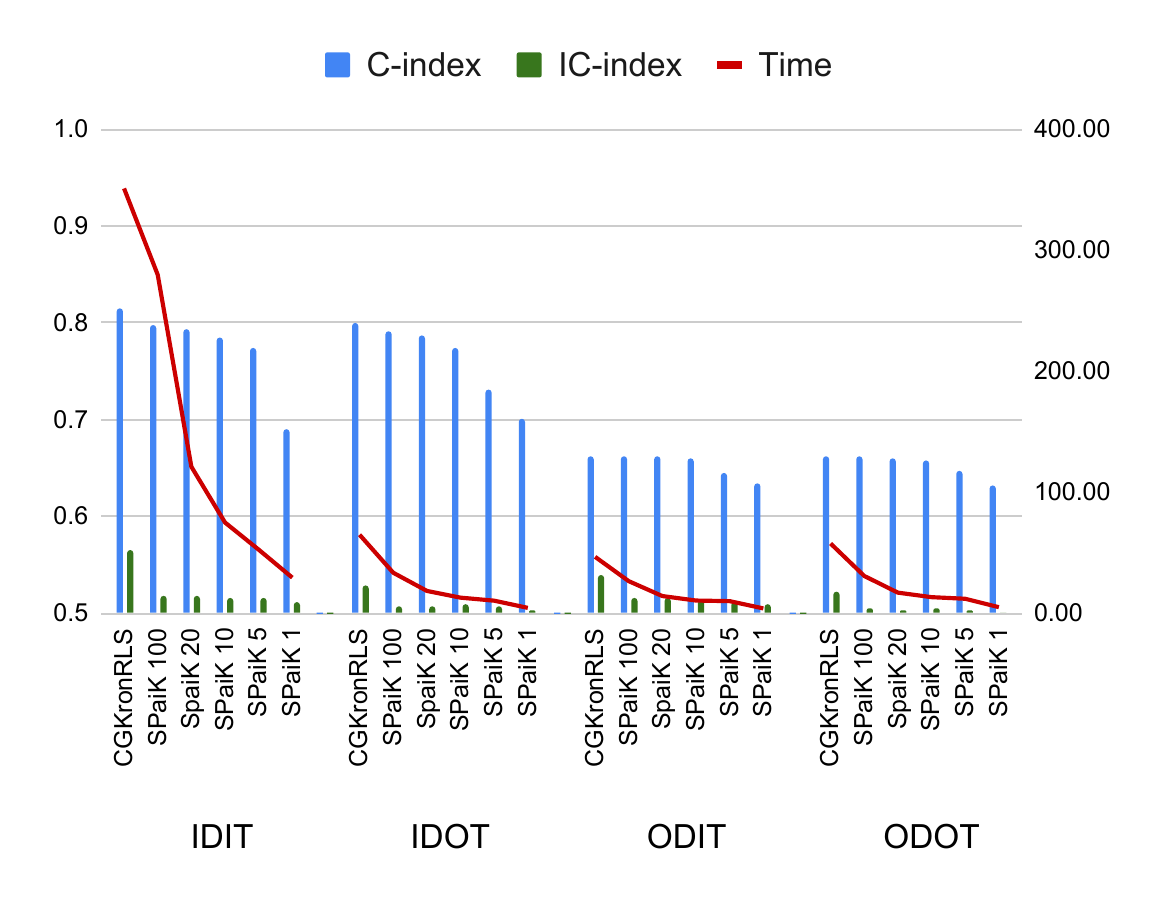}}
\caption{C-index (blue bar), IC-index (green bar), and time (red line) of pairwise kernel learning algorithms in data with continuous labels under different experimental settings.} 
\label{fig1}
\end{figure} 

\begin{figure} [ht]  
\vspace{-2ex}  
 %\centering 
  \subfigure[GPCR]{
   \includegraphics[width=0.50\textwidth]{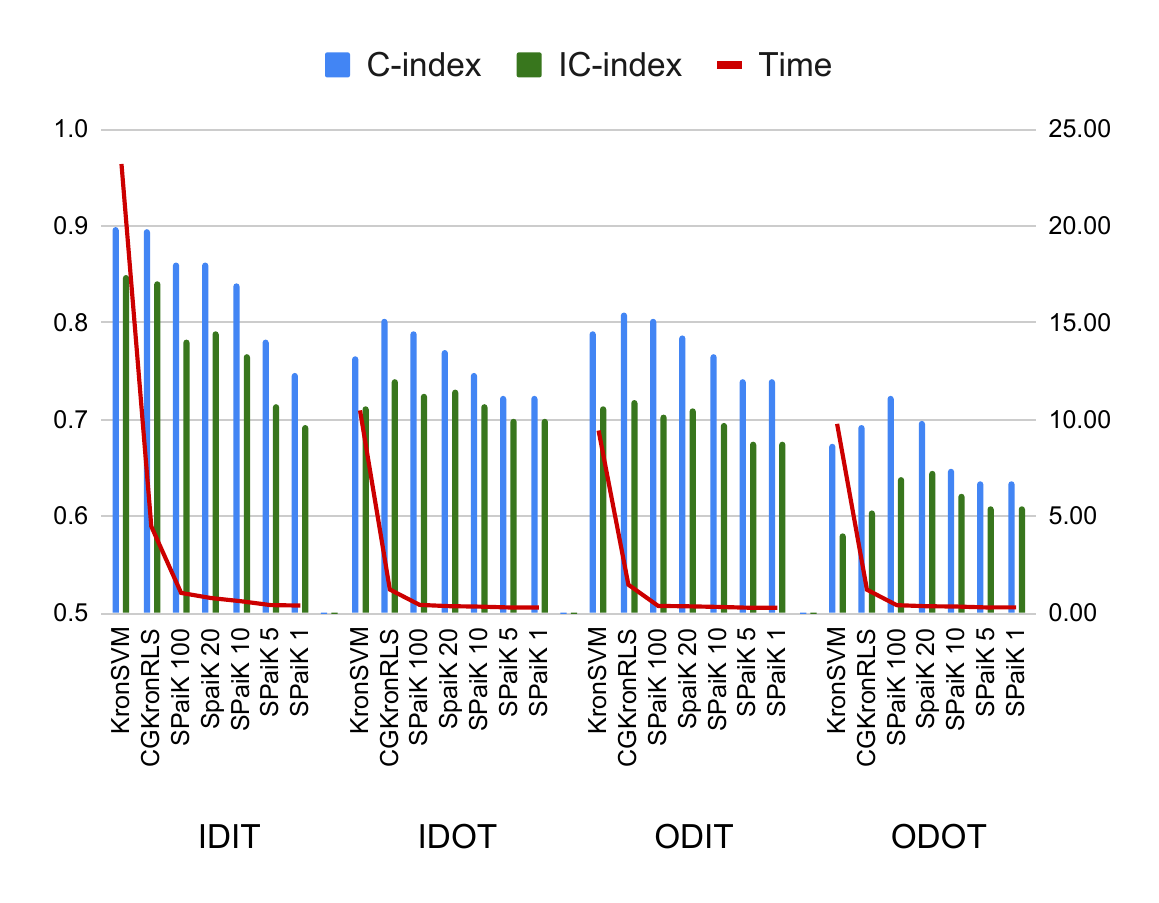}}%\hspace{-3.7em}
  \subfigure[Ion Channels]{
   \includegraphics[width=0.50\textwidth]  {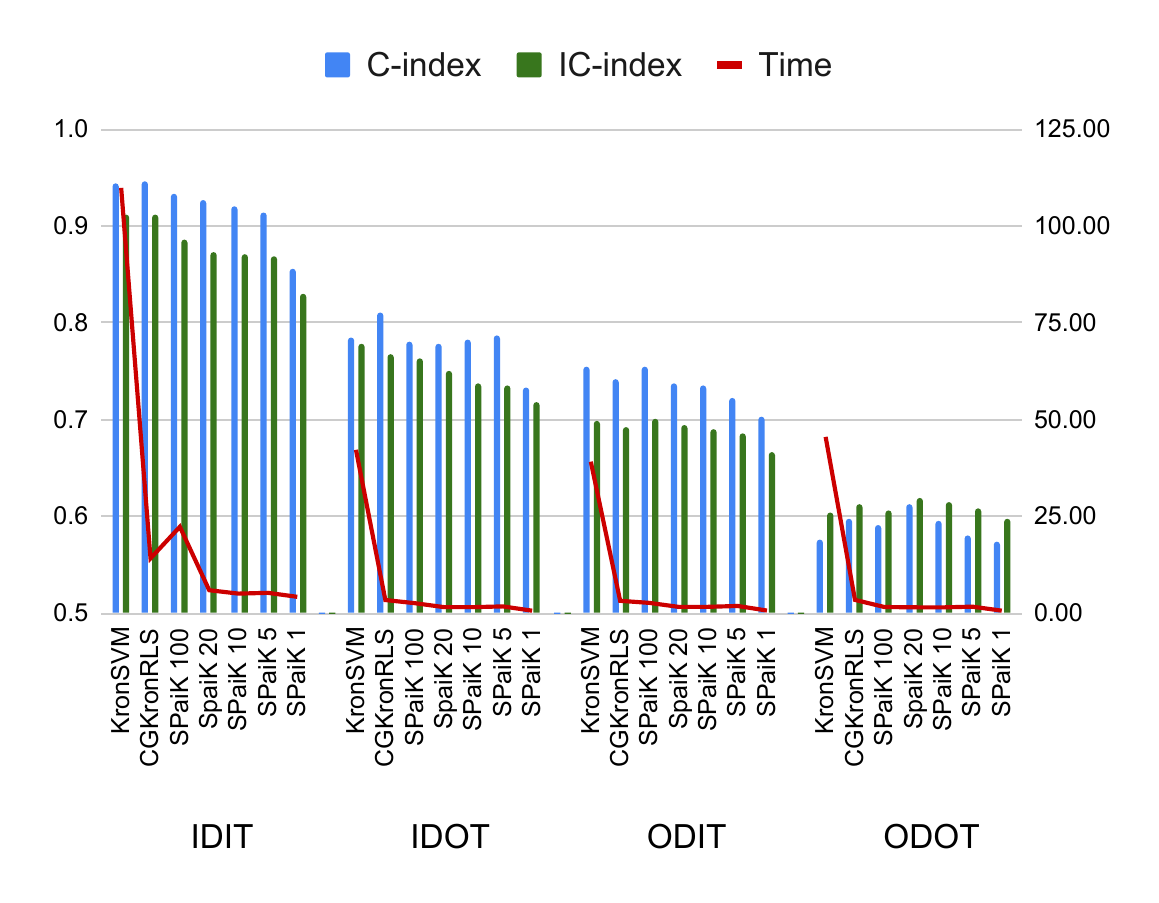}}\\%\hspace{-3.7em}
  \subfigure[Enzymes]{\label{fig2c}
   \includegraphics[width=0.50\textwidth]  {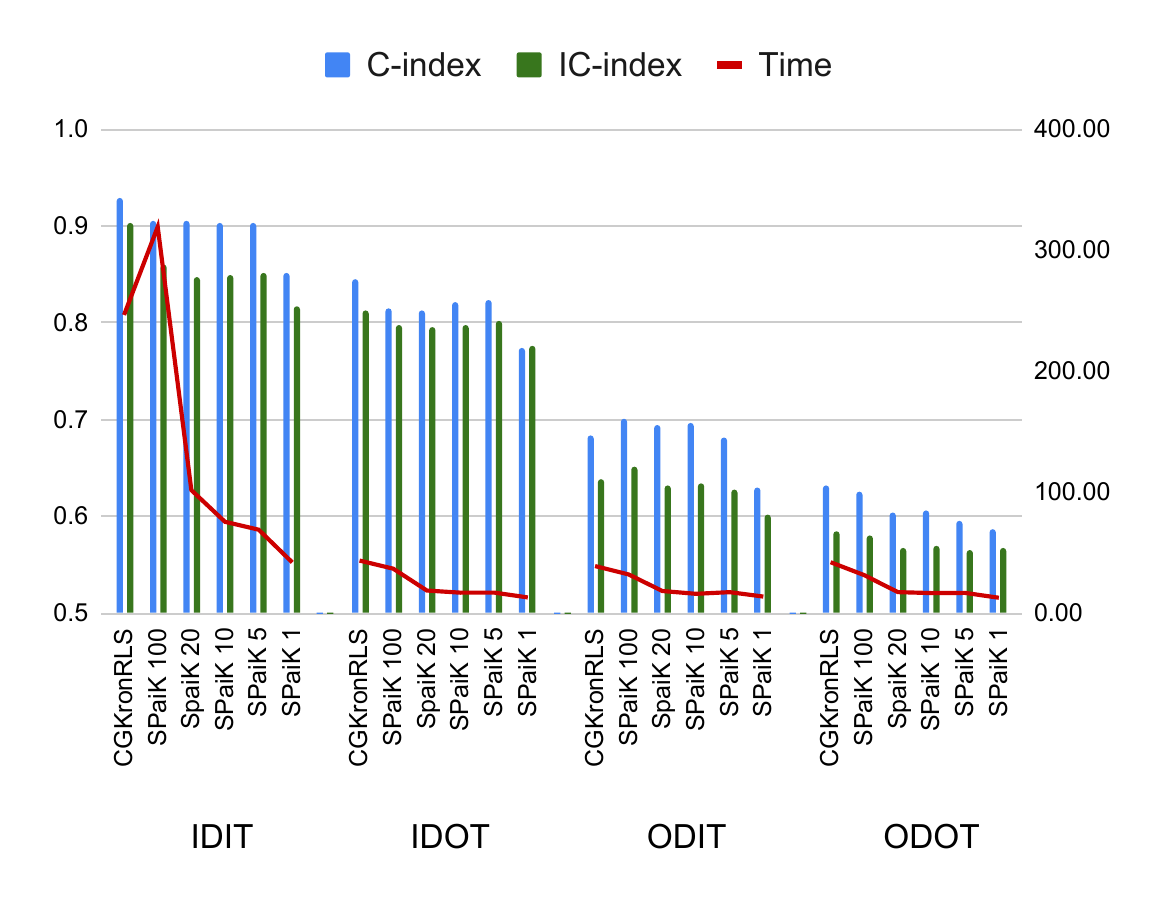}}%\hspace{-3.7em}
  %\subfigure[Merget]{
  % \includegraphics[width=0.50\textwidth]{{metz}}\label{cis4}}
\caption{C-index (blue bar), IC-index (green bar), and time (red line) of pairwise kernel learning algorithms in data with binary labels under different experimental settings. In the Enzymes dataset, KronSVM was excluded from experiments because it failed to complete the training even on the first IDIT data within five hours.
} 
\label{fig2}
\end{figure}

\subsubsection{General Observations} % Muutettu
As expected, the best prediction accuracy was achieved on the most informative IDIT data, whereas the more realistic IDOT, ODIT, and ODOT settings resulted in lower accuracies (see Figures~\ref{fig1} and \ref{fig2}). In many cases, results on the IDOT data outperformed those on the ODIT data, indicating that predicting new targets for known drugs tends to be easier than predicting new drugs for known targets. The most difficult scenario for all methods was zero-shot learning (ODOT). Nevertheless, even in this challenging setting, a predictive model could be successfully trained on most datasets. In addition, we note that prediction is generally easier in binary-labeled datasets (GPCR, Ion Channels, and Enzymes) than in continuously labeled datasets (Davis, Metz, KiBA, and Merget), especially so on IDIT data.  Finally, we remark that the longer computational times observed on IDIT data are explained by its larger size, which leads to a correspondingly larger problem to be solved. These observations are consistent with earlier findings in \cite{Kar_pwl_2025,pauliina,pahikkala2,AIpaper2025}.

%\subsubsection{Comparison of {\tt SPaiK} with {\tt CGKronRLS} and {\tt KronSVM}} 
\subsubsection{Comparison of {\tt SPaiK100} with {\tt CGKronRLS} and {\tt KronSVM}} 

{\tt KronSVM} was evaluated only on the binary-labeled datasets, where it proved to be by far the most time-consuming method --- on average about 6 and 10 times slower than {\tt CGKronRLS} on GPCR and Ion Channels, respectively --- and did not finish within 5 hours on the Enzymes dataset (whereas {\tt CGKronRLS} required only about 370 seconds (see Figure \ref{fig2}). Although {\tt KronSVM} occasionally produced more accurate results than {\tt CGKronRLS}, this was neither consistent nor the typical outcome. For this reason, we consider {\tt CGKronRLS} the more relevant baseline for evaluating {\tt SPaiK100}. We recall that {\tt SPaiK100} corresponds to using the (non-stochastic) GVT. The differences between {\tt CGKronRLS} and {\tt SPaiK100} lie in the choice of loss function (least squares vs.\ $\varepsilon$-insensitive squared loss), regularization (Euclidean norm vs.\ $\ell_1$-norm), and optimization method (CG vs.\ InexactLMBM). Moreover, {\tt CGKronRLS} is a well-established and mature implementation, whereas {\tt SPaiK} represents a novel combination of algorithms. These distinctions help to explain the performance differences observed between the two methods, while also highlighting that {\tt SPaiK} achieves competitive results despite its novelty.
%These distinctions help to explain the performance differences observed between the two methods in our experiments.

A quick look at Figures \ref{fig1} and \ref{fig2}
reveals that {\tt CGKronRLS} is the most accurate predictor on the simplest IDIT data and often also on IDOT and ODIT data. Nevertheless, on the latter two the gap narrows and {\tt SPaiK100} matches or nearly matches {\tt CGKronRLS} across datasets. Interestingly, on Davis, Ion Channels, and Enzymes under the ODIT setting, {\tt SPaiK100} even surpasses {\tt CGKronRLS}. 

%(e.g., Davis 0.794 vs 0.797; Metz 0.676 vs 0.681; KiBA 0.698 vs 0.705; Merget 0.793 vs 0.800). \spkl20 remains close; deeper stochasticity (
%≤5) starts to degrade.
%In zero-shot learning (S4), {\tt SPaiK100} produced more accurate predictions than {\tt CGKronRLS} in all but one dataset (Enzymes). This improvement is likely due to the use of the nonsmooth optimization solver \LMBMNS, which is less sensitive to perturbations than the smooth CG solver. A similar behavior was observed in \cite{Kar_pwl_2025}.
%\enlargethispage{\baselineskip}
% tässä menossa numeeristen tulosten IDIT vaihdoksia

In zero-shot learning (ODOT), {\tt SPaiK100} produced more accurate predictions than {\tt CGKronRLS} in all but two dataset: Ion Channels and Enzymes. This improvement is likely related to the differences in optimization and modeling choices: {\tt SPaiK100} employs the nonsmooth optimization solver \iLMBMNS, which is less sensitive to perturbations than the smooth CG solver, together with an $\varepsilon$-insensitive squared loss and $\ell_1$-regularization. These choices may provide additional robustness in the zero-shot setting, where both drugs and targets are unseen during training and the learning task is inherently more difficult. A similar behavior was observed in \cite{Kar_pwl_2025} with sparse pairwise kernel learning models.

The CPU times of {\tt SPaiK100} --- the full-batch version of {\tt SPaiK} --- are mostly comparable to, or shorter than, those of {\tt CGKronRLS}. Longer CPU times for {\tt SPaiK100} are observed only for the KiBA, Ion Channels, and Enzymes datasets under the IDIT setting. Since in {\tt SPaiK} the batch size $n_\batch$ (or, equivalently, the batch percentage $p_\batch$) directly controls the computational cost, we next examine how varying this parameter affects both runtime and predictive accuracy. 
 
\subsubsection{Different Batch Sizes}
Across both continuously labeled and binary labeled datasets, the experiments show a clear accuracy--efficiency trade-off in {\tt SPaiK}. As the batch percentage $p_\batch$ decreases, the computational cost is substantially reduced, while predictive accuracy, measured by the C-index and IC-index, gradually decreases. Overall, the most favorable trade-off is obtained with moderate batch percentages, especially around $p_\batch=20$. In contrast, very small batches ($p_\batch=1$--$5$) give the shortest CPU times, but they also lead to more noticeable accuracy losses and are therefore mainly suitable for cases where runtime is the dominant constraint.

For the continuous datasets, reducing the batch percentage from $p_\batch=100$ to $p_\batch=20$ has only a minor effect on predictive accuracy. The average decrease in the C-index value from {\tt SPaiK100} to {\tt SPaiK20} is $0.003$, indicating that {\tt SPaiK20} maintains almost the same predictive performance as the full-batch version. 
Further reductions in the batch percentage lead to larger accuracy losses: for {\tt SPaiK10}, {\tt SPaiK5}, and {\tt SPaiK1}, the average decreases are $0.011$, $0.030$, and $0.059$.
 A similar trend is observed for the binary datasets, where the corresponding decreases in the C-index values are $0.008$, $0.018$, $0.033$, and $0.061$ for {\tt SPaiK20}, {\tt SPaiK10}, {\tt SPaiK5}, and {\tt SPaiK1}, respectively. Thus, moderate batch percentages preserve predictive accuracy well, whereas the smallest batch percentages lead to a more visible degradation.

At the same time, the reduction in CPU time is substantial. In particular, {\tt SPaiK20} is considerably faster than {\tt SPaiK100}, sometimes by an order of magnitude, while maintaining nearly the same accuracy. This makes {\tt SPaiK20} a robust default choice and a particularly attractive compromise between predictive performance and computational cost. In the more realistic IDOT, ODIT, and ODOT settings, {\tt SPaiK20} often provides the best overall trade-off. This is particularly evident in the zero-shot setting (ODOT), where {\tt SPaiK20} is competitive with, and in some cases better than, the baseline methods, indicating that stochastic batching does not weaken generalization in the most challenging setting.

The IC-index further supports the effectiveness of \sgvtns. Even with relatively small batch percentages ($p_\batch \geq 10$), \sgvt is able to capture pairwise interactions between drugs and targets. This is partly due to the auxiliary dual--drug matrix $M$, which preserves information from previous iterations of the \sgvt algorithm and thereby maintains dependencies across batches. A notable exception is the Merget dataset, where the IC-index values are lower across all variants of {\tt SPaiK}, including {\tt SPaiK100} (see Figure \ref{fig1d}). This suggests that the weaker performance on Merget is more likely related to the properties of the dataset itself than to stochastic batching.

Overall, the results indicate that {\tt SPaiK20} offers substantial computational gains with little loss in accuracy. Although {\tt CGKronRLS} often gives highly accurate results, especially in the more informative IDIT setting, {\tt SPaiK20} provides a more scalable alternative with a strong accuracy--efficiency balance. Smaller batch percentages ($p_\batch \leq 5$) can be useful when computational efficiency is the main priority, but they should be used cautiously due to the increased loss in predictive accuracy.
Finally, we note that \sgvt is not tied to \spkl or the \slmbm optimizer: as a stochastic matrix--vector multiplication technique for Kronecker-product kernels, it could in principle also be incorporated into other Kronecker-product kernel methods, including CG-based RLS approaches such as {\tt CGKronRLS}.

\subsubsection{Comparison of Epoch-Wise and Random Batch Selection}\label{sec_epoch_vs_random}
\begin{figure} [ht]  
\vspace{-2ex}  
 %\centering 
  \subfigure[C-index]{
   \includegraphics[width=0.30\textwidth]{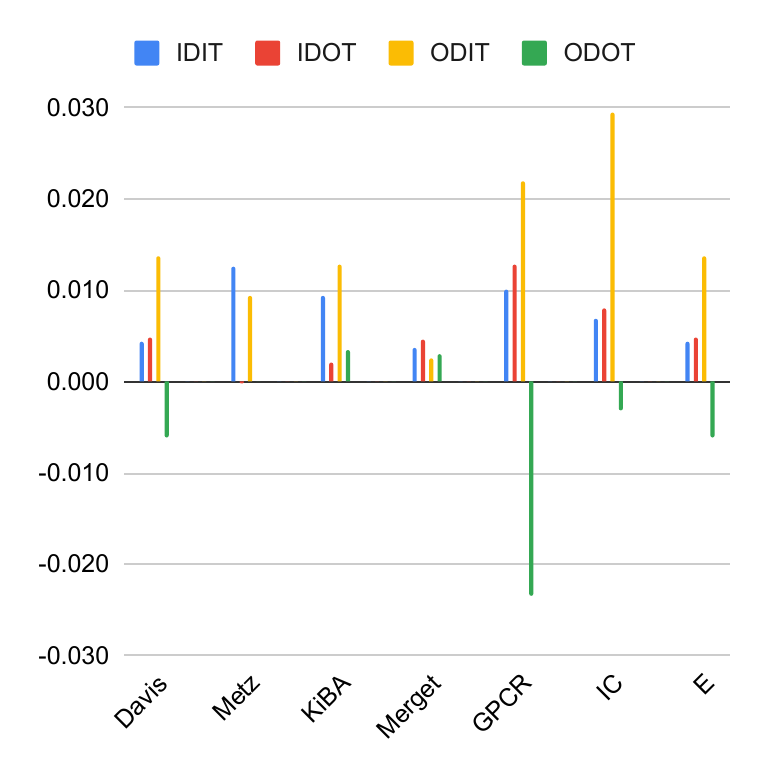}}%\hspace{-3.7em}
  \subfigure[IC-index]{
   \includegraphics[width=0.30\textwidth]  {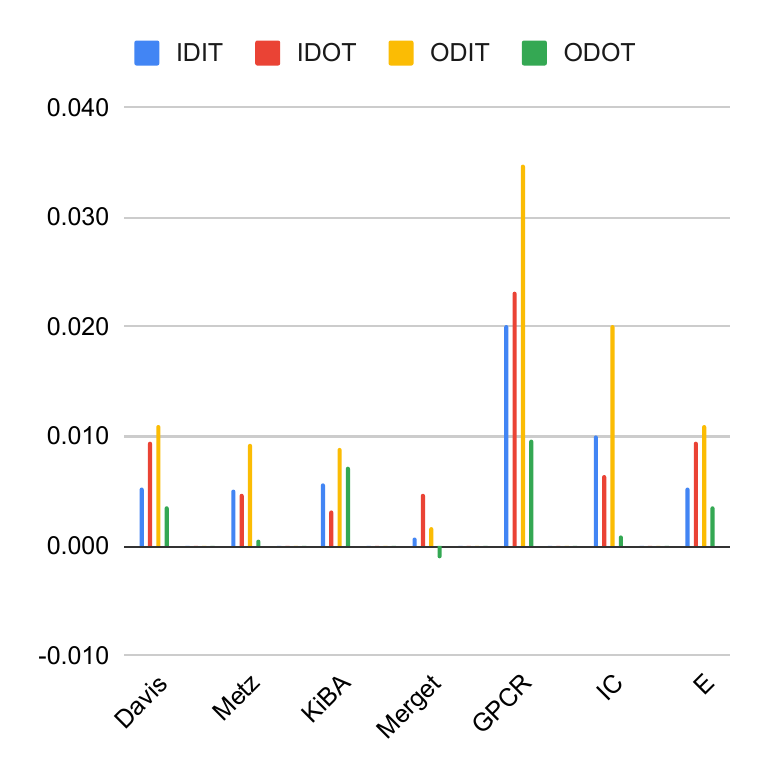}}%\hspace{-3.7em}
  \subfigure[CPU-time]{\label{fig3c}
   \includegraphics[width=0.30\textwidth]  {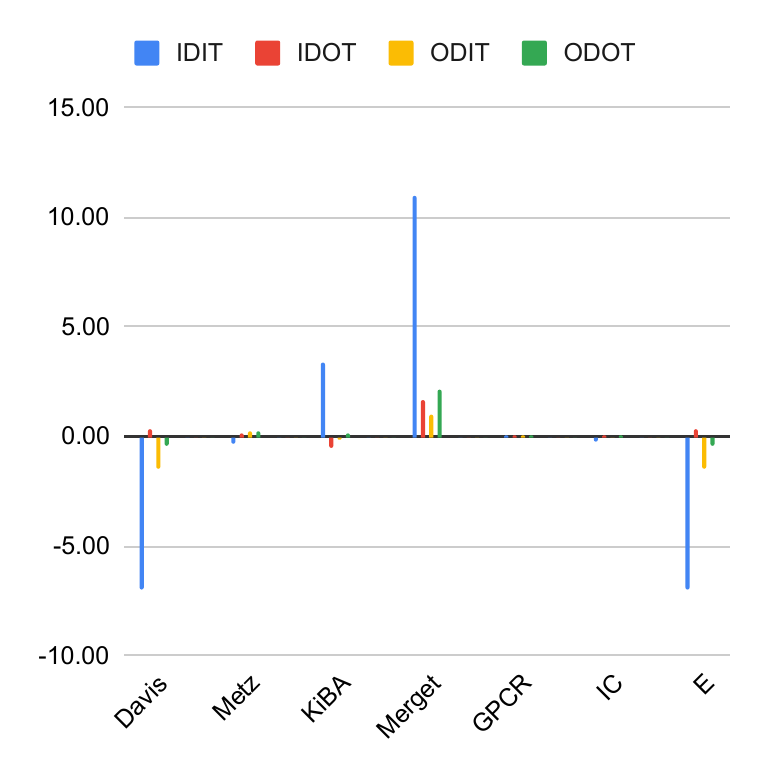}}%\hspace{-3.7em}
  %\subfigure[Merget]{
  % \includegraphics[width=0.50\textwidth]{{metz}}\label{cis4}}
\caption{Comparison of epoch-wise and fully random target-wise batch selection for {\tt SPaiK}10. The figures show the differences in (a) C-index, (b) IC-index, and (c) CPU-time between {\tt SPaiK}10 and {\tt SPaiK-R}10, where positive values indicate better performance of the epoch-wise variant {\tt SPaiK}10.} 
\label{fig3}
\end{figure}

Finally, to assess the effect of the epoch-wise batch selection, we compare {\tt SPaiK}10 with a fully random target-wise variant, denoted by {\tt SPaiK-R}10. The two variants use the same batch size, but differ in how the target indices are selected: {\tt SPaiK}10 processes the target indices epoch-wise in random order, whereas {\tt SPaiK-R}10 selects the target indices independently at random for each batch. We compare the two variants in terms of C-index, IC-index, and CPU time over all seven datasets in Figure \ref{fig3}.

In general, the C-index is usually higher for the epoch-wise variant than for the fully random variant, although on the ODOT datasets the fully random variant occasionally obtains better values. An interesting observation, however, is that the IC-index, which measures pairwise interaction-level performance, is almost always better for the epoch-wise variant. The only exception is the Merget dataset in the ODOT setting, where the values are very close; however, for Merget the IC-index values are close to 0.5 (the trivial case) also in general, see Figure~\ref{fig1d}. This suggests that the epoch-wise batch selection is particularly beneficial for preserving pairwise interaction information, even in cases where the C-index alone would indicate better performance for the fully random variant. This is also natural, since the epoch-wise approach approximates the full Kronecker-product kernel matrix more systematically, whereas with fully random target-wise batch selection some parts of the interaction structure may remain insufficiently represented.

The CPU times of the two variants are generally comparable. The largest differences are observed for some IDIT datasets, where the difference reaches up to approximately $15$\% of the total computational time. Since the differences occur in both directions, neither variant shows a systematic computational advantage.

\section{Conclusions}
In this work, we introduced the {\em stochastic generalized vec trick} (sGVT), a stochastic extension of the generalized vec trick (GVT) for efficient multiplication of Kronecker-product kernel matrices with vectors. 
From the naive case to our stochastic method, the computational cost of this multiplication improves as follows:
\begin{itemize}
    \item \textbf{Naive computation:} $\mathcal{O}(n^2)$, where $n$ is the number of drug-target pairs.
    \item \textbf{GVT:} $\mathcal{O}(nm + nq)$, where $m,q \ll n$ are numbers of unique drugs and targets, respectively; 
    \item \textbf{sGVT:} $\mathcal{O}(n_\batch m + n_\batch q)$, where $n_\batch \ll n$ is the batch size defined by the user.
\end{itemize}

Building on \sgvtns, we further developed \spklns, a scalable pairwise kernel learning algorithm that combines  this efficient kernel computation with a new stochastic optimizer, the stochastic inexact limited-memory bundle method (\slmbmns). We emphasize, however, that \sgvt is not restricted to \spklns: as a stochastic kernel matrix--vector multiplication technique, it can also be combined with other optimization and learning methods that rely on Kronecker-product kernel computations.

We evaluated \spkl on seven real-world drug--target affinity datasets and compared it against established kernel baselines, {\tt CGKronRLS} and {\tt KronSVM}. The results show that \spkl is highly competitive with these state-of-the-art methods. Although {\tt CGKronRLS} gives the highest accuracy in the most informative IDIT setting, \spkl performs strongly in the more challenging experimental settings. In particular, in the zero-shot learning (ODOT), where both drugs and targets are unseen during training, \spkl remains competitive with the baselines and outperforms them in several cases.

Among the \spkl variants, the full-batch version, {\tt SPaiK100}, provides a useful reference point, while {\tt SPaiK20} achieves nearly the same accuracy with substantially lower computational cost. This indicates that stochastic batching can preserve predictive performance while making kernel-based pairwise learning considerably more scalable. Moreover, the consistently high interaction concordance index (IC-index) values show that the stochastic version of GVT can successfully capture pairwise interactions between drugs and targets\footnote{The only exception was the Merget dataset, where performance was lower across all methods, including \spkl with full batch, suggesting that the limitation stems from the dataset rather than from stochastic batching.}.

Overall, the experiments demonstrate that \spkl is a practical and scalable alternative for both continuous and binary pairwise learning tasks. A batch percentage of around $p_\batch=20$ provides the best balance between runtime and predictive accuracy, extending the applicability of pairwise kernel methods to problem sizes that are difficult to handle with existing full-batch approaches.

%Tämä on täällä, koska kannattanee mainita tuo operaatioiden määrän väheneminen täällä conclusioneissa
%
%However, efficient matrix–vector multiplication can be achieved using the {\em generalized vec trick} (GVT) \cite{airola,viljanen}, a sparse Kronecker product algorithm that reduces the cost of such operations from quadratic ($\mathcal{O}(n^2)$, where $n$ is the number of observed pairs) to a form that scales essentially linearly with the number of drugs and targets ($\mathcal{O}(nm+nq)$, where $m$ and $q$ are the numbers of unique drugs and targets, respectively).
%This breakthrough has enabled pairwise kernel methods to scale to much larger datasets than was previously possible, since typically $m,q \ll n$.
\section*{Acknowledgements}
The work was financially supported by the Research Council of Finland, Projects No.\ \#340182 and \#345804 led by  Tapio Pahikkala and Projects No.\ \#340140 and \#345805 led by Antti Airola.

The authors acknowledge the use of ChatGPT by OpenAI for language editing. The authors take full responsibility for the content of the manuscript.
%#340182 ja #340140 
%\newpage
%\bibliographystyle{alpha}
\bibliographystyle{acm}
%\bibliography{references}

\newpage
\section*{Appendix}
The results of our numerical experiments are given in Tables~\ref{tab:davis}--\ref{tab:e}. 
We recall that the results are averaged over five different training, validation, and test set splits of each data. In addition, within the different data splits, the stochastic method \spkl is executed with (and averaged over) five different random batch selections.

\begin{table}[ht]
    \caption{Results in Davis.}
    \centering 
    %\resizebox{1.00\textwidth}{!}{
    \begin{tabular}{llrrrrrrrrrrrrrrrrrrrrrrr} \hline
        Setting & Method && C-index && IC-index && MSE &&  \multicolumn{1}{c}{CPU} \\ 
        \hline
        IDIT	&	CGKronRLS	&	&	0.845	&&	0.692	&&	0.395	&&	10.67	\\
	&	SPaiK 100	&	&	0.822	&&	0.639	&&	0.474	&&	8.40	\\
	%&	Spaik-TR 20	&	&	0.814	&&	0.630	&&	0.496	&&	4.74	\\
	&	SpaiK 20	&	&	0.813	&&	0.628	&&	0.498	&&	4.39	\\
	&	SpaiK-R 10	&	&	0.797	&&	0.626	&&	0.535	&&	2.66	\\
	&	SPaiK 10	&	&	0.808	&&	0.631	&&	0.509	&&	2.91	\\
	%&	SpaiK-TR 5	&	&	0.791	&&	0.615	&&	0.558	&&	2.19	\\
	&	SPaiK 5	    &	&	0.798	&&	0.621	&&	0.544	&&	2.32	\\
	%&	SpaiK-TR 1	&	&	0.758	&&	0.597	&&	0.677	&&	1.51	\\
	&	SPaiK 1	    &	&	0.763	&&	0.601	&&	0.680	&&	1.64	\\
													\hline	
IDOT	&	CGKronRLS	&	&	0.797	&&	0.649	&&	0.577	&&	2.05	\\
	&	SPaiK 100	&	&	0.794	&&	0.638	&&	0.558	&&	1.42	\\
	%&	Spaik-TR 20	&	&	0.787	&&	0.630	&&	0.584	&&	0.80	\\
	&	SpaiK 20	&	&	0.790	&&	0.633	&&	0.575	&&	0.89	\\
	&	SpaiK-R 10	&	&	0.780	&&	0.625	&&	0.624	&&	0.72	\\
	&	SPaiK 10	&	&	0.782	&&	0.630	&&	0.613	&&	0.76	\\
	%&	SpaiK-TR 5	&	&	0.771	&&	0.619	&&	0.652	&&	0.65	\\
	&	SPaiK 5	    &	&	0.777	&&	0.625	&&	0.653	&&	0.74	\\
	%&	SpaiK-TR 1	&	&	0.752	&&	0.607	&&	0.777	&&	0.51	\\
	&	SPaiK 1	    &	&	0.737	&&	0.613	&&	0.839	&&	0.57	\\
												\hline												
ODIT	&	CGKronRLS	&	&	0.678	&&	0.572	&&	0.827	&&	2.15	\\
	&	SPaiK 100	&	&	0.679	&&	0.564	&&	0.826	&&	1.26	\\
	%&	Spaik-TR 20	&	&	0.672	&&	0.560	&&	0.869	&&	0.79	\\
	&	SpaiK 20	&	&	0.675	&&	0.560	&&	0.856	&&	0.79	\\
	&	SpaiK-R 10	&	&	0.660	&&	0.555	&&	0.898	&&	0.71	\\
	&	SPaiK 10	&	&	0.666	&&	0.559	&&	0.866	&&	0.76	\\
	%&	SpaiK-TR 5	&	&	0.656	&&	0.555	&&	0.912	&&	0.66	\\
	&	SPaiK 5	    &	&	0.658	&&	0.556	&&	0.917	&&	0.69	\\
	%&	SpaiK-TR 1	&	&	0.623	&&	0.546	&&	1.072	&&	0.49	\\
	&	SPaiK 1	    &	&	0.619	&&	0.547	&&	1.119	&&	0.53	\\
	\hline																					
ODOT	&	CGKronRLS	&	&	0.614	&&	0.550	&&	1.005	&&	2.16	\\
	&	SPaiK 100	&	&	0.634	&&	0.550	&&	0.956	&&	1.27	\\
	%&	Spaik-TR 20	&	&	0.638	&&	0.548	&&	0.988	&&	0.77	\\
	&	SpaiK 20	&	&	0.638	&&	0.548	&&	0.981	&&	0.82	\\
	&	SpaiK-R 10	&	&	0.628	&&	0.546	&&	1.009	&&	0.71	\\
	&	SPaiK 10	&	&	0.634	&&	0.547	&&	1.017	&&	0.73	\\
	%&	SpaiK-TR 5	&	&	0.630	&&	0.544	&&	1.038	&&	0.65	\\
	&	SPaiK 5	    &	&	0.627	&&	0.543	&&	1.042	&&	0.69	\\
	%&	SpaiK-TR 1	&	&	0.622	&&	0.535	&&	1.155	&&	0.50	\\
	&	SPaiK 1 	&	&	0.606	&&	0.539	&&	1.203	&&	0.56	\\
\hline															
    \end{tabular}%}
    \label{tab:davis}
\end{table}

\begin{table}[ht]
    \caption{Results in Metz.}
    \centering 
    %\resizebox{1.00\textwidth}{!}{
    \begin{tabular}{llrrrrrrrrrrrrrrrrrrrrrrr} \hline
        Setting & Method && C-index && IC-index && MSE &&  \multicolumn{1}{c}{CPU} \\ 
        \hline
        IDIT	&	CGKronRLS	&	&	0.799	&&	0.680	&&	0.221	&&	101.22	\\
	&	SPaiK 100	&	&	0.763	&&	0.614	&&	0.280	&&	80.42	\\
	%&	Spaik-TR 20	&	&	0.755	&&	0.612	&&	0.293	&&	42.75	\\
	&	SpaiK 20	&	&	0.761	&&	0.617	&&	0.282	&&	41.46	\\
	&	SpaiK-R 10	&	&	0.727	&&	0.610	&&	0.335	&&	24.53	\\
	&	SPaiK 10	&	&	0.740	&&	0.615	&&	0.316	&&	24.23	\\
	%&	SpaiK-TR 5	&	&	0.696	&&	0.606	&&	0.389	&&	17.24	\\
	&	SPaiK 5 	&	&	0.710	&&	0.613	&&	0.367	&&	17.87	\\
	%&	SpaiK-TR 1	&	&	0.657	&&	0.589	&&	0.517	&&	8.24	\\
	&	SPaiK 1 	&	&	0.656	&&	0.595	&&	0.632	&&	8.61	\\
												\hline	
IDOT	&	CGKronRLS	&	&	0.681	&&	0.595	&&	0.381	&&	16.81	\\
	&	SPaiK 100	&	&	0.674	&&	0.575	&&	0.395	&&	10.02	\\
	%&	Spaik-TR 20	&	&	0.673	&&	0.569	&&	0.409	&&	5.22	\\
	&	SpaiK 20	&	&	0.675	&&	0.573	&&	0.419	&&	5.47	\\
	&	SpaiK-R 10	&	&	0.669	&&	0.569	&&	0.428	&&	4.76	\\
	&	SPaiK 10	&	&	0.669	&&	0.574	&&	0.424	&&	4.86	\\
	%&	SpaiK-TR 5	&	&	0.652	&&	0.568	&&	0.493	&&	3.73	\\
	&	SPaiK 5 	&	&	0.644	&&	0.571	&&	0.541	&&	3.97	\\
	%&	SpaiK-TR 1	&	&	0.652	&&	0.568	&&	0.493	&&	3.76	\\
	&	SPaiK 1 	&	&	0.644	&&	0.571	&&	0.541	&&	4.02	\\
												\hline											
ODIT	&	CGKronRLS	&	&	0.700	&&	0.644	&&	0.436	&&	16.98	\\
	&	SPaiK 100	&	&	0.687	&&	0.603	&&	0.453	&&	9.53	\\
	%&	Spaik-TR 20	&	&	0.676	&&	0.602	&&	0.480	&&	6.87	\\
	&	SpaiK 20	&	&	0.680	&&	0.605	&&	0.476	&&	6.78	\\
	&	SpaiK-R 10	&	&	0.653	&&	0.592	&&	0.554	&&	4.92	\\
	&	SPaiK 10	&	&	0.662	&&	0.602	&&	0.521	&&	5.19	\\
	%&	SpaiK-TR 5	&	&	0.617	&&	0.578	&&	0.654	&&	3.68	\\
	&	SPaiK 5 	&	&	0.614	&&	0.583	&&	0.728	&&	3.99	\\
	%&	SpaiK-TR 1	&	&	0.617	&&	0.578	&&	0.654	&&	3.67	\\
	&	SPaiK 1 	&	&	0.614	&&	0.583	&&	0.728	&&	4.03	\\
												\hline									
ODOT	&	CGKronRLS	&	&	0.601	&&	0.549	&&	0.610	&&	13.77	\\
	&	SPaiK 100	&	&	0.603	&&	0.553	&&	0.576	&&	11.97	\\
	%&	Spaik-TR 20	&	&	0.607	&&	0.550	&&	0.602	&&	5.54	\\
	&	SpaiK 20	&	&	0.608	&&	0.551	&&	0.623	&&	5.41	\\
	&	SpaiK-R 10	&	&	0.607	&&	0.549	&&	0.626	&&	4.67	\\
	&	SPaiK 10	&	&	0.607	&&	0.550	&&	0.649	&&	4.91	\\
	%&	SpaiK-TR 5	&	&	0.600	&&	0.548	&&	0.675	&&	3.92	\\
	&	SPaiK 5 	&	&	0.594	&&	0.549	&&	0.694	&&	4.07	\\
	%&	SpaiK-TR 1	&	&	0.600	&&	0.548	&&	0.675	&&	3.91	\\
	&	SPaiK 1 	&	&	0.594	&&	0.549	&&	0.694	&&	4.15	\\
\hline															
    \end{tabular}%}
    \label{tab:metz}
\end{table}

\begin{table}[ht]
    \caption{Results in KiBA.}
    \centering 
    %\resizebox{1.00\textwidth}{!}{
    \begin{tabular}{llrrrrrrrrrrrrrrrrrrrrrrr} \hline
        Setting & Method && C-index && IC-index && MSE &&  \multicolumn{1}{c}{CPU} \\ 
        \hline
        IDIT	&	CGKronRLS	&	&	0.805	&&	0.690	&&	0.273	&&	218.59	\\
	&	SPaiK 100	&	&	0.775	&&	0.644	&&	0.344	&&	261.10	\\
	%&	Spaik-TR 20	&	&	0.774	&&	0.645	&&	0.347	&&	111.78	\\
	&	SpaiK 20	&	&	0.774	&&	0.645	&&	0.346	&&	103.33	\\
	&	SpaiK-R 10	&	&	0.757	&&	0.642	&&	0.373	&&	62.55	\\
	&	SPaiK 10	&	&	0.767	&&	0.648	&&	0.354	&&	65.94	\\
	%&	SpaiK-TR 5	&	&	0.730	&&	0.630	&&	0.437	&&	41.77	\\
	&	SPaiK 5 	&	&	0.745	&&	0.641	&&	0.404	&&	43.33	\\
	%&	SpaiK-TR 1	&	&	0.649	&&	0.597	&&	0.898	&&	20.38	\\
	&	SPaiK 1 	&	&	0.644	&&	0.600	&&	0.989	&&	20.56	\\
												\hline									
IDOT	&	CGKronRLS	&	&	0.705	&&	0.634	&&	0.468	&&	32.62	\\
	&	SPaiK 100	&	&	0.692	&&	0.608	&&	0.506	&&	36.07	\\
	%&	Spaik-TR 20	&	&	0.688	&&	0.604	&&	0.528	&&	14.28	\\
	&	SpaiK 20	&	&	0.692	&&	0.607	&&	0.520	&&	13.25	\\
	&	SpaiK-R 10	&	&	0.679	&&	0.595	&&	0.598	&&	9.62	\\
	&	SPaiK 10	&	&	0.682	&&	0.598	&&	0.596	&&	9.14	\\
	%&	SpaiK-TR 5	&	&	0.660	&&	0.581	&&	0.780	&&	8.26	\\
	&	SPaiK 5 	&	&	0.663	&&	0.586	&&	0.717	&&	8.20	\\
	%&	SpaiK-TR 1	&	&	0.647	&&	0.571	&&	0.975	&&	2.91	\\
	&	SPaiK 1 	&	&	0.646	&&	0.580	&&	1.295	&&	2.90	\\
												\hline									
ODIT	&	CGKronRLS	&	&	0.680	&&	0.660	&&	1.612	&&	26.71	\\
	&	SPaiK 100	&	&	0.672	&&	0.635	&&	1.633	&&	37.33	\\
	%&	Spaik-TR 20	&	&	0.651	&&	0.626	&&	1.692	&&	16.35	\\
	&	SpaiK 20	&	&	0.661	&&	0.632	&&	1.675	&&	15.63	\\
	&	SpaiK-R 10	&	&	0.633	&&	0.618	&&	1.774	&&	12.74	\\
	&	SPaiK 10	&	&	0.646	&&	0.627	&&	1.751	&&	12.57	\\
	%&	SpaiK-TR 5	&	&	0.610	&&	0.599	&&	2.049	&&	10.73	\\
	&	SPaiK 5     &	&	0.623	&&	0.613	&&	1.955	&&	11.34	\\
	%&	SpaiK-TR 1	&	&	0.599	&&	0.590	&&	2.114	&&	3.91	\\
	&	SPaiK 1 	&	&	0.601	&&	0.599	&&	2.231	&&	3.86	\\
												\hline									
ODOT	&	CGKronRLS	&	&	0.602	&&	0.581	&&	1.897	&&	21.15	\\
	&	SPaiK 100	&	&	0.603	&&	0.582	&&	1.824	&&	40.26	\\
	%&	Spaik-TR 20	&	&	0.597	&&	0.573	&&	1.871	&&	15.66	\\
	&	SpaiK 20	&	&	0.600	&&	0.577	&&	1.835	&&	16.29	\\
	&	SpaiK-R 10	&	&	0.597	&&	0.566	&&	1.914	&&	12.32	\\
	&	SPaiK 10	&	&	0.600	&&	0.573	&&	1.890	&&	12.49	\\
	%&	SpaiK-TR 5	&	&	0.588	&&	0.561	&&	2.035	&&	11.35	\\
	&	SPaiK 5 	&	&	0.590	&&	0.564	&&	2.050	&&	10.99	\\
	%&	SpaiK-TR 1	&	&	0.581	&&	0.557	&&	2.210	&&	3.90	\\
	&	SPaiK 1 	&	&	0.582	&&	0.562	&&	2.704	&&	4.00	\\
\hline															
    \end{tabular}%}
    \label{tab:kiba}
\end{table}

\begin{table}[ht]
    \caption{Results in Merget.}
    \centering 
    %\resizebox{1.00\textwidth}{!}{
    \begin{tabular}{llrrrrrrrrrrrrrrrrrrrrrrr} \hline
        Setting & Method && C-index && IC-index && MSE &&  \multicolumn{1}{c}{CPU} \\ 
        \hline
        IDIT	&	CGKronRLS	&	&	0.816	&&	0.565	&&	0.303	&&	351.13	\\
	&	SPaiK 100	&	&	0.797	&&	0.517	&&	0.345	&&	279.81	\\
	%&	Spaik-TR 20	&	&	0.789	&&	0.516	&&	0.356	&&	109.61	\\
	&	SpaiK 20	&	&	0.793	&&	0.517	&&	0.349	&&	121.11	\\
	&	SpaiK-R 10	&	&	0.782	&&	0.516	&&	0.370	&&	64.00	\\
	&	SPaiK 10	&	&	0.786	&&	0.517	&&	0.359	&&	75.01	\\
	%&	SpaiK-TR 5	&	&	0.771	&&	0.514	&&	0.386	&&	50.32	\\
	&	SPaiK 5 	&	&	0.773	&&	0.517	&&	0.378	&&	52.58	\\
	%&	SpaiK-TR 1	&	&	0.693	&&	0.514	&&	0.582	&&	28.42	\\
	&	SPaiK 1 	&	&	0.689	&&	0.512	&&	0.672	&&	29.44	\\
\hline																											
IDOT	&	CGKronRLS	&	&	0.800	&&	0.528	&&	0.342	&&	64.88	\\
	&	SPaiK 100	&	&	0.791	&&	0.508	&&	0.364	&&	33.48	\\
	%&	Spaik-TR 20	&	&	0.785	&&	0.504	&&	0.381	&&	16.20	\\
	&	SpaiK 20	&	&	0.787	&&	0.507	&&	0.373	&&	18.46	\\
	&	SpaiK-R 10	&	&	0.769	&&	0.504	&&	0.411	&&	11.24	\\
	&	SPaiK 10	&	&	0.774	&&	0.509	&&	0.401	&&	12.88	\\
	%&	SpaiK-TR 5	&	&	0.738	&&	0.505	&&	0.552	&&	11.03	\\
	&	SPaiK 5 	&	&	0.731	&&	0.506	&&	0.602	&&	10.38	\\
	%&	SpaiK-TR 1	&	&	0.710	&&	0.504	&&	0.564	&&	4.12	\\
	&	SPaiK 1 	&	&	0.701	&&	0.503	&&	0.660	&&	4.43	\\
\hline																											
ODIT	&	CGKronRLS	&	&	0.662	&&	0.539	&&	0.670	&&	46.62	\\
	&	SPaiK 100	&	&	0.662	&&	0.516	&&	0.636	&&	26.51	\\
	%&	Spaik-TR 20	&	&	0.659	&&	0.514	&&	0.654	&&	12.47	\\
	&	SpaiK 20	&	&	0.662	&&	0.516	&&	0.643	&&	14.19	\\
	&	SpaiK-R 10	&	&	0.657	&&	0.511	&&	0.683	&&	9.51	\\
	&	SPaiK 10	&	&	0.660	&&	0.513	&&	0.686	&&	10.48	\\
	%&	SpaiK-TR 5	&	&	0.647	&&	0.513	&&	0.741	&&	10.35	\\
	&	SPaiK 5 	&	&	0.645	&&	0.513	&&	0.740	&&	9.98	\\
	%&	SpaiK-TR 1	&	&	0.637	&&	0.509	&&	0.766	&&	3.98	\\
	&	SPaiK 1 	&	&	0.635	&&	0.509	&&	0.760	&&	3.98	\\
\hline																																
ODOT	&	CGKronRLS	&	&	0.663	&&	0.523	&&	0.660	&&	57.61	\\
	&	SPaiK 100	&	&	0.663	&&	0.506	&&	0.650	&&	30.80	\\
	%&	Spaik-TR 20	&	&	0.659	&&	0.506	&&	0.670	&&	14.43	\\
	&	SpaiK 20	&	&	0.660	&&	0.503	&&	0.672	&&	17.01	\\
	&	SpaiK-R 10	&	&	0.654	&&	0.506	&&	0.688	&&	11.12	\\
	&	SPaiK 10	&	&	0.657	&&	0.505	&&	0.709	&&	13.28	\\
	%&	SpaiK-TR 5	&	&	0.645	&&	0.503	&&	0.766	&&	11.77	\\
	&	SPaiK 5 	&	&	0.647	&&	0.504	&&	0.814	&&	11.95	\\
	%&	SpaiK-TR 1	&	&	0.637	&&	0.506	&&	0.823	&&	4.71	\\
	&	SPaiK 1 	&	&	0.631	&&	0.501	&&	0.852	&&	5.00	\\
\hline															
    \end{tabular}%}
    \label{tab:merget}
\end{table}

\begin{table}[ht]
    \caption{Results in GPCR.}
    \centering 
    %\resizebox{1.00\textwidth}{!}{
    \begin{tabular}{llrrrrrrrrrrrrrrrrrrrrrrr} \hline
        Setting & Method && C-index && IC-index && MSE &&  \multicolumn{1}{c}{CPU} \\ 
        \hline
        IDIT	&	KronSVM	&	&	0.899	&&	0.849	&&	0.024	&&	23.21	\\
	&	CGKronRLS	&	&	0.896	&&	0.842	&&	0.024	&&	4.49	\\
	&	SPaiK 100	&	&	0.862	&&	0.783	&&	0.027	&&	1.05	\\
	%&	Spaik-TR 20	&	&	0.850	&&	0.767	&&	0.028	&&	0.70	\\
	&	SpaiK 20	&	&	0.863	&&	0.791	&&	0.030	&&	0.79	\\
	&	SpaiK-R 10	&	&	0.832	&&	0.748	&&	0.030	&&	0.60	\\
	&	SPaiK 10	&	&	0.842	&&	0.769	&&	0.030	&&	0.62	\\
	%&	SpaiK-TR 5	&	&	0.770	&&	0.696	&&	0.034	&&	0.41	\\
	&	SPaiK 5 	&	&	0.782	&&	0.715	&&	0.034	&&	0.42	\\
	%&	SpaiK-TR 1	&	&	0.733	&&	0.670	&&	0.033	&&	0.35	\\
	&	SPaiK 1 	&	&	0.748	&&	0.695	&&	0.036	&&	0.40	\\
				\hline								
IDOT	&	KronSVM	&	&	0.765	&&	0.715	&&	0.025	&&	10.48	\\
	&	CGKronRLS	&	&	0.805	&&	0.741	&&	0.028	&&	1.22	\\
	&	SPaiK 100	&	&	0.792	&&	0.726	&&	0.029	&&	0.43	\\
	%&	Spaik-TR 20	&	&	0.770	&&	0.718	&&	0.031	&&	0.35	\\
	&	SpaiK 20	&	&	0.773	&&	0.731	&&	0.032	&&	0.36	\\
	&	SpaiK-R 10	&	&	0.736	&&	0.693	&&	0.036	&&	0.33	\\
	&	SPaiK 10	&	&	0.749	&&	0.716	&&	0.036	&&	0.34	\\
	%&	SpaiK-TR 5	&	&	0.714	&&	0.683	&&	0.038	&&	0.29	\\
	&	SPaiK 5   	&	&	0.724	&&	0.701	&&	0.042	&&	0.30	\\
	%&	SpaiK-TR 1	&	&	0.714	&&	0.683	&&	0.038	&&	0.29	\\
	&	SPaiK 1	    &	&	0.724	&&	0.701	&&	0.042	&&	0.29	\\
			\hline									
ODIT	&	KronSVM	&	&	0.792	&&	0.715	&&	0.029	&&	9.44	\\
	&	CGKronRLS	&	&	0.811	&&	0.721	&&	0.032	&&	1.47	\\
	&	SPaiK 100	&	&	0.805	&&	0.705	&&	0.030	&&	0.38	\\
	%&	Spaik-TR 20	&	&	0.769	&&	0.690	&&	0.032	&&	0.35	\\
	&	SpaiK 20	&	&	0.788	&&	0.711	&&	0.032	&&	0.36	\\
	&	SpaiK-R 10	&	&	0.747	&&	0.661	&&	0.033	&&	0.30	\\
	&	SPaiK 10	&	&	0.769	&&	0.696	&&	0.034	&&	0.33	\\
	%&	SpaiK-TR 5	&	&	0.738	&&	0.672	&&	0.034	&&	0.28	\\
	&	SPaiK 5	    &	&	0.742	&&	0.677	&&	0.034	&&	0.29	\\
	%&	SpaiK-TR 1	&	&	0.738	&&	0.672	&&	0.034	&&	0.27	\\
	&	SPaiK 1	    &	&	0.742	&&	0.677	&&	0.034	&&	0.29	\\
		\hline										
ODOT	&	KronSVM	&	&	0.675	&&	0.582	&&	0.031	&&	9.78	\\
	&	CGKronRLS	&	&	0.693	&&	0.607	&&	0.034	&&	1.22	\\
	&	SPaiK 100	&	&	0.724	&&	0.640	&&	0.033	&&	0.42	\\
	%&	Spaik-TR 20	&	&	0.700	&&	0.632	&&	0.035	&&	0.36	\\
	&	SpaiK 20	&	&	0.699	&&	0.648	&&	0.036	&&	0.36	\\
	&	SpaiK-R 10	&	&	0.673	&&	0.613	&&	0.036	&&	0.33	\\
	&	SPaiK 10	&	&	0.650	&&	0.623	&&	0.039	&&	0.35	\\
	%&	SpaiK-TR 5	&	&	0.633	&&	0.608	&&	0.038	&&	0.29	\\
	&	SPaiK 5	    &	&	0.636	&&	0.610	&&	0.041	&&	0.30	\\
	%&	SpaiK-TR 1	&	&	0.633	&&	0.608	&&	0.038	&&	0.28	\\
	&	SPaiK 1	    &	&	0.636	&&	0.610	&&	0.041	&&	0.30	\\
	\hline														
    \end{tabular}%}
    \label{tab:gpcr}
\end{table}

\begin{table}[ht]
    \caption{Results in Ion Channels.}
    \centering 
    %\resizebox{1.00\textwidth}{!}{
    \begin{tabular}{llrrrrrrrrrrrrrrrrrrrrrrr} \hline
        Setting & Method && C-index && IC-index && MSE &&  \multicolumn{1}{c}{CPU} \\ 
        \hline
IDIT	&	KronSVM	&	&	0.945	&&	0.913	&&	0.019	&&	109.88	\\
	&	CGKronRLS	&	&	0.946	&&	0.912	&&	0.019	&&	14.18	\\
	&	SPaiK 100	&	&	0.934	&&	0.886	&&	0.021	&&	22.33	\\
	%&	Spaik-TR 20	&	&	0.925	&&	0.872	&&	0.023	&&	6.69	\\
	&	SpaiK 20	&	&	0.927	&&	0.873	&&	0.023	&&	5.94	\\
	&	SpaiK-R 10	&	&	0.915	&&	0.860	&&	0.024	&&	5.25	\\
	&	SPaiK 10	&	&	0.922	&&	0.870	&&	0.023	&&	5.07	\\
	%&	SpaiK-TR 5	&	&	0.899	&&	0.845	&&	0.025	&&	5.13	\\
	&	SPaiK 5	&	&	0.915	&&	0.869	&&	0.025	&&	5.27	\\
	%&	SpaiK-TR 1	&	&	0.856	&&	0.816	&&	0.031	&&	4.33	\\
	&	SPaiK 1	&	&	0.855	&&	0.829	&&	0.035	&&	4.24	\\
		\hline											
IDOT	&	KronSVM	&	&	0.784	&&	0.779	&&	0.023	&&	42.21	\\
	&	CGKronRLS	&	&	0.810	&&	0.768	&&	0.024	&&	3.41	\\
	&	SPaiK 100	&	&	0.781	&&	0.762	&&	0.025	&&	2.60	\\
	%&	Spaik-TR 20	&	&	0.780	&&	0.740	&&	0.026	&&	1.54	\\
	&	SpaiK 20	&	&	0.779	&&	0.749	&&	0.026	&&	1.57	\\
	&	SpaiK-R 10	&	&	0.776	&&	0.731	&&	0.027	&&	1.50	\\
	&	SPaiK 10	&	&	0.784	&&	0.738	&&	0.026	&&	1.57	\\
	%&	SpaiK-TR 5	&	&	0.751	&&	0.720	&&	0.028	&&	1.69	\\
	&	SPaiK 5	    &	&	0.787	&&	0.736	&&	0.027	&&	1.76	\\
	%&	SpaiK-TR 1	&	&	0.722	&&	0.704	&&	0.030	&&	0.66	\\
	&	SPaiK 1	    &	&	0.734	&&	0.718	&&	0.032	&&	0.66	\\
		\hline											
ODIT	&	KronSVM	&	&	0.755	&&	0.698	&&	0.034	&&	39.15	\\
	&	CGKronRLS	&	&	0.742	&&	0.692	&&	0.035	&&	3.21	\\
	&	SPaiK 100	&	&	0.755	&&	0.701	&&	0.034	&&	2.67	\\
	%&	Spaik-TR 20	&	&	0.733	&&	0.687	&&	0.035	&&	1.68	\\
	&	SpaiK 20	&	&	0.737	&&	0.695	&&	0.035	&&	1.63	\\
	&	SpaiK-R 10	&	&	0.705	&&	0.669	&&	0.036	&&	1.73	\\
	&	SPaiK 10	&	&	0.734	&&	0.689	&&	0.037	&&	1.65	\\
	%&	SpaiK-TR 5	&	&	0.689	&&	0.664	&&	0.037	&&	1.82	\\
	&	SPaiK 5	    &	&	0.722	&&	0.686	&&	0.038	&&	1.88	\\
	%&	SpaiK-TR 1	&	&	0.653	&&	0.656	&&	0.038	&&	0.67	\\
	&	SPaiK 1	    &	&	0.703	&&	0.666	&&	0.037	&&	0.69	\\
\hline													
ODOT	&	KronSVM	&	&	0.576	&&	0.604	&&	0.036	&&	45.55	\\
	&	CGKronRLS	&	&	0.598	&&	0.614	&&	0.036	&&	3.45	\\
	&	SPaiK 100	&	&	0.592	&&	0.606	&&	0.035	&&	1.61	\\
	%&	Spaik-TR 20	&	&	0.606	&&	0.623	&&	0.036	&&	1.52	\\
	&	SpaiK 20	&	&	0.613	&&	0.619	&&	0.036	&&	1.53	\\
	&	SpaiK-R 10	&	&	0.598	&&	0.613	&&	0.036	&&	1.45	\\
	&	SPaiK 10	&	&	0.595	&&	0.614	&&	0.037	&&	1.51	\\
	%&	SpaiK-TR 5	&	&	0.578	&&	0.607	&&	0.038	&&	1.61	\\
	&	SPaiK 5	    &	&	0.580	&&	0.608	&&	0.038	&&	1.70	\\
	%&	SpaiK-TR 1	&	&	0.595	&&	0.596	&&	0.038	&&	0.66	\\
	&	SPaiK 1	    &	&	0.574	&&	0.597	&&	0.039	&&	0.66	\\
\hline															
    \end{tabular}%}
    \label{tab:ic}
\end{table}

\begin{table}[ht]
    \caption{ Results in Enzymes.}
    \centering 
    %\resizebox{1.00\textwidth}{!}{
    \begin{tabular}{llrrrrrrrrrrrrrrrrrrrrrrr} \hline
        Setting & Method && C-index && IC-index && MSE &&  \multicolumn{1}{c}{CPU} \\ 
        \hline
        IDIT	&	CGKronRLS	&	&	0.928	&	&	0.904	&	&	0.005	&	&	246.57	\\
	&	SPaiK 100	&	&	0.906	&	&	0.860	&	&	0.006	&	&	319.41	\\
	%&	Spaik-TR 20	&	&	0.908	&	&	0.850	&	&	0.007	&	&	105.89	\\
	&	SpaiK 20	&	&	0.906	&	&	0.848	&	&	0.007	&	&	101.78	\\
	&	SpaiK-R 10	&	&	0.900	&	&	0.843	&	&	0.007	&	&	82.57	\\
	&	SPaiK 10	&	&	0.904	&	&	0.849	&	&	0.007	&	&	75.58	\\
	%&	SpaiK-TR 5	&	&	0.897	&	&	0.842	&	&	0.007	&	&	68.37	\\
	&	SPaiK 5	&	&	0.903	&	&	0.852	&	&	0.007	&	&	69.06	\\
	%&	SpaiK-TR 1	&	&	0.842	&	&	0.808	&	&	0.008	&	&	40.30	\\
	&	SPaiK 1	&	&	0.851	&	&	0.818	&	&	0.008	&	&	42.02	\\
			\hline												
IDOT	&	CGKronRLS	&	&	0.844	&	&	0.813	&	&	0.005	&	&	43.52	\\
	&	SPaiK 100	&	&	0.816	&	&	0.798	&	&	0.006	&	&	36.82	\\
	%&	Spaik-TR 20	&	&	0.816	&	&	0.790	&	&	0.006	&	&	18.78	\\
	&	SpaiK 20	&	&	0.814	&	&	0.796	&	&	0.006	&	&	18.76	\\
	&	SpaiK-R 10	&	&	0.817	&	&	0.789	&	&	0.006	&	&	16.68	\\
	&	SPaiK 10	&	&	0.822	&	&	0.798	&	&	0.006	&	&	17.00	\\
	%&	SpaiK-TR 5	&	&	0.814	&	&	0.787	&	&	0.006	&	&	16.99	\\
	&	SPaiK 5	&	&	0.824	&	&	0.802	&	&	0.006	&	&	17.06	\\
	%&	SpaiK-TR 1	&	&	0.765	&	&	0.763	&	&	0.008	&	&	12.45	\\
	&	SPaiK 1	&	&	0.774	&	&	0.776	&	&	0.008	&	&	13.05	\\
		\hline													
ODIT	&	CGKronRLS	&	&	0.683	&	&	0.640	&	&	0.009	&	&	38.99	\\
	&	SPaiK 100	&	&	0.700	&	&	0.652	&	&	0.008	&	&	32.04	\\
	%&	Spaik-TR 20	&	&	0.685	&	&	0.626	&	&	0.008	&	&	18.77	\\
	&	SpaiK 20	&	&	0.694	&	&	0.632	&	&	0.008	&	&	18.35	\\
	&	SpaiK-R 10	&	&	0.682	&	&	0.623	&	&	0.008	&	&	17.53	\\
	&	SPaiK 10	&	&	0.696	&	&	0.634	&	&	0.008	&	&	16.05	\\
	%&	SpaiK-TR 5	&	&	0.672	&	&	0.621	&	&	0.008	&	&	17.07	\\
	&	SPaiK 5	&	&	0.682	&	&	0.628	&	&	0.009	&	&	17.43	\\
	%&	SpaiK-TR 1	&	&	0.651	&	&	0.612	&	&	0.009	&	&	13.64	\\
	&	SPaiK 1	&	&	0.630	&	&	0.602	&	&	0.010	&	&	13.76	\\
	\hline														
ODOT	&	CGKronRLS	&	&	0.632	&	&	0.586	&	&	0.010	&	&	42.18	\\
	&	SPaiK 100	&	&	0.625	&	&	0.580	&	&	0.008	&	&	31.48	\\
	%&	Spaik-TR 20	&	&	0.606	&	&	0.569	&	&	0.009	&	&	18.08	\\
	&	SpaiK 20	&	&	0.604	&	&	0.568	&	&	0.009	&	&	17.45	\\
	&	SpaiK-R 10	&	&	0.613	&	&	0.565	&	&	0.009	&	&	17.05	\\
	&	SPaiK 10	&	&	0.607	&	&	0.569	&	&	0.009	&	&	16.65	\\
	%&	SpaiK-TR 5	&	&	0.607	&	&	0.561	&	&	0.009	&	&	17.39	\\
	&	SPaiK 5	&	&	0.595	&	&	0.565	&	&	0.009	&	&	16.80	\\
	%&	SpaiK-TR 1	&	&	0.594	&	&	0.562	&	&	0.010	&	&	12.86	\\
	&	SPaiK 1	&	&	0.587	&	&	0.567	&	&	0.010	&	&	12.79	\\
	\hline									
    \end{tabular}%}
    \label{tab:e}
\end{table}

\end{document}